\documentclass[journal]{IEEEtran}
\usepackage{graphicx}
\usepackage{subcaption}
\usepackage{comment}
\usepackage{amsmath}

\usepackage{color}
\usepackage[dvipsnames]{xcolor}

\usepackage[switch]{lineno}
\ifCLASSINFOpdf
\else
\fi
\begin{document}
%
\title{High-Resolution Semantic Labeling \\
       with Convolutional Neural Networks}
%
%
%

\author{Emmanuel~Maggiori,~\IEEEmembership{Student member,~IEEE,}
        Yuliya~Tarabalka,~\IEEEmembership{Member,~IEEE,} \\
        Guillaume~Charpiat,
        and~Pierre~Alliez

\thanks{E.~Maggiori, Y.~Tarabalka and P.~Alliez are with  Universit\'e C\^ote d'Azur, TITANE team, Inria, 2004 route des Lucioles, BP93 06902 Sophia Antipolis Cedex, France. E-mail: emmanuel.maggiori@inria.fr.}
\thanks{G.~Charpiat is with Tao team, Inria Saclay--\^Ile-de-France, Laboratoire de Recherche en Informatique, Universit\'e Paris-Sud, 91405 Orsay Cedex, France.}

\thanks{Manuscript received ...; revised ...}}

%
%

\markboth{High-Resolution Semantic Labeling
       with Convolutional Neural Networks}%
{Shell \MakeLowercase{\textit{et al.}}: Bare Demo of IEEEtran.cls for IEEE Journals}
%

\maketitle


\begin{abstract}
Convolutional neural networks (CNNs) have received increasing attention over the last few years. They were initially conceived for image categorization, i.e., the problem of assigning a semantic label to an entire input image.

In this paper we address the problem of dense semantic labeling, which consists in assigning a semantic label to \emph{every} pixel in an image. Since this requires a high spatial accuracy to determine \emph{where} labels are assigned,  categorization CNNs, intended to be highly robust to local deformations, are not directly applicable.

By adapting categorization networks, many semantic labeling CNNs have been recently proposed. Our first contribution is an in-depth analysis of these architectures. We establish the desired properties of an ideal semantic labeling CNN, and assess how those methods stand with regard to these properties. We observe that even though they provide competitive results, these CNNs often underexploit properties of semantic labeling that could lead to more effective and efficient architectures.

Out of these observations, we then derive a CNN framework specifically adapted to the semantic labeling problem. In addition to learning features at different resolutions, it learns how to combine these features. By integrating local and global information in an efficient and flexible manner, it outperforms previous techniques. We evaluate the proposed framework and compare it with state-of-the-art architectures on public benchmarks of high-resolution aerial image labeling.

\end{abstract}


%
\IEEEpeerreviewmaketitle

\section{Introduction}


\IEEEPARstart{S}{emantic} labeling is the problem of assigning a semantic class to every individual pixel of an image. In certain application domains, such as high-resolution aerial image analysis, it is of paramount importance to provide fine-grained classification maps where object boundaries are precisely located. For example, with the advent of autonomous driving there is an increasing interest in locating the exact boundaries of roads or even lanes from aerial imagery~\cite{urtasun_lanes}.

Over the last few years, \emph{deep learning} and, in particular, convolutional neural networks (CNNs), have gained significant attention in the image analysis community. These have been originally devised for the image \emph{categorization} problem, i.e.,~the assignment of one label to an entire image. For example, they have been used to categorize objects in natural scenes (e.g., \emph{airplane}, \emph{bird}, \emph{person}) or land use in the case of aerial images (e.g., \emph{forest}, \emph{beach}, \emph{tennis court}). CNNs jointly learn to extract relevant contextual features and conduct the categorization. In addition to the suppression of the feature design process, which is an interesting advantage itself, this technique has consistently beaten alternative methods in a wide range of problems~\cite{visualizingcnn}. Nowadays, one can reasonably expect to find CNN-based techniques scoring the best positions in the leaderboards of online image-related contests.

While neural networks have existed for a few decades, a combination of recent advances has facilitated their development as deep learning techniques. One of these advances is the use of novel architectures. Notably, the novelty in the aforementioned convolutional network is its architecture, which imposes significant restrictions to the neuronal connections compared to previous approaches. While CNNs are thus less general than traditional architectures, the  restrictions applied are well grounded in the domain of image analysis, reducing thus the optimization search space in a sensible way. This directs the network to learn a more appropriate function, yielding better categorization results. The lesson learned is that finding the right type of architecture for a given problem often boosts the performance of neural networks. Moreover, fewer computational resources are required for training and conducting labeling.

A sort of ``recipe'' or meta-architecture for the image categorization problem 
was incrementally developed in the community. The typical ingredients of a state-of-the-art CNN to categorize images are a combination of convolutional, so-called pooling layers and rectifed linear units, followed by traditional fully-connected layers. However, when it comes to semantic pixel labeling (i.e., assigning a class to every pixel), this categorization recipe cannot be directly transferred. 
Indeed,
while categorization networks are devised to lose spatial precision in order to identify objects that come in different appearances, semantic labeling networks should preserve the spatial resolution to correctly locate object boundaries. This is not straightforward 
to implement,
because of a well-known trade-off between recognition and localization~\cite{fcn,deeplab},
due to the need to keep the networks small (and thus more efficient and easier to train).
Since both qualities are required in semantic labeling at the same time, it is important to design specific architectures for this problem.

There have been recent research efforts in this direction, using CNNs 
 for pixel labeling and, in particular, for high-resolution aerial image labeling (e.g.,~\cite{volpi,sherrah}). These networks certainly provide good results and stand as competitive alternatives compared to other methods. However, 
there is still a need for finding 
\emph{optimal} architectures for semantic labeling, i.e., the ``recipe'' for 
suitable semantic labeling networks. 
We consider that just by doing a proper analysis of the architecture 
required for our problem we may develop smaller, more efficient networks to achieve equivalent or even better results.

Our first contribution is a detailed analysis of the main families of CNN architectures proposed 
recently
for the semantic labeling problem. We group the different methods into three categories: \emph{dilation} (e.g., \cite{sherrah,mammography}), \emph{deconvolution} (e.g., \cite{unpooling,segnet1,realtimenet,volpi}) and  \emph{skip} (e.g., \cite{fcn,isprs_ensemble}) networks.
These categories are different 
from
each other in the way of addressing the aforementioned recognition/localization trade-off. For example, while the  networks by Long et al.~\cite{fcn} and Marmanisa et al.~\cite{isprs_ensemble} are substantially different in structure and application domain, they are both \emph{skip} networks in how they manage to provide a high-resolution output. After establishing the desired properties of a semantic labeling architecture, we position the different families of networks with respect to these properties. 
Let us remark that it is also common to include post-processing modules to increase the resolution of CNN's outputs, such as fully connected CRFs \cite{sherrah,deeplab,isprs_effective}. 
However, our review focuses on architectures that are specifically designed to provide a high-resolution output.


Our second contribution is a novel semantic labeling network architecture, referred to as MLP (after \emph{multi-layer perceptron}).
 Derived from the notion of \emph{skip} network, the MLP architecture yields high flexibility and expressiveness by extracting features at different resolutions (and thus at different levels of details), and \emph{learning} how to combine them in order to generate fine-grained classification maps. In the literature, probably the most similar approach is the one in~\cite{hypercolumns} which, though 
for
a different problem, also seeks to learn to combine multi-resolution features.
Our MLP architecture exhibits a better performance in aerial image labeling than many other recent techniques, despite being simpler and smaller than them. 
The design of an appropriate architecture thus leads to a win-win situation, in which both accuracy and computational complexity are improved.


We conduct experiments on two popular benchmarks for high-resolution aerial segmentation: Vaihingen and Potsdam datasets, proposed as part of the ISPRS Semantic Labeling Contest~\cite{2dSemLabeling}. These datasets highlight the specific challenges of aerial image labeling, requiring to outline small objects with a high spatial precision.


This paper first introduces convolutional neural networks and their use in semantic labeling (Sec.~\ref{s:cnns}). An analysis of the different high-resolution labeling schemes  is then presented (Sec.~\ref{s:analysis}). We later describe our proposed architecture (Sec.~\ref{s:mlp}) and perform an experimental evaluation (Sec.~\ref{s:experiments}). Conclusions are drawn in Sec.~\ref{s:concl}.

\section{Convolutional Neural Networks}
\label{s:cnns}

An artificial neural network is a system of interconnected neurons that pass messages to each other. When the messages are passed from one neuron to the next 
one
without 
ever
going back (i.e., the graph of message passing is acyclic) they network is referred to as feed-forward \cite{neuralNetsBook}, which is the most 
common
type of network 
in
image categorization. 
An individual neuron takes a vector of inputs $\mathbf{x} = x_1\dots x_n$ and performs a simple operation to produce an output $a$. The most common neuron is defined as follows:
\begin{equation}
a = \sigma ( \mathbf{w} \mathbf{x} + b ),
\label{eq:neuron}
\end{equation}
where $\mathbf{w}$ denotes a weight vector, $b$ a scalar known as \emph{bias} and $\sigma$ an activation function. Put simply, a neuron computes a weighted sum of its inputs and applies a possibly nonlinear scalar function on the result. The weights $\mathbf{w}$ and biases $b$ are the parameters of the neurons that define the function. The goal of training is to find the optimal values for these parameters,
so that the function computed by the neural network performs the best on the task assigned. 

The most common activation functions $\sigma$ are sigmoids, hyperbolic tangents and rectified linear units (ReLU). 
For image analysis, ReLUs have become the most popular choice due to some practical advantages at training time, but novel activation functions have been recently proposed as well~\cite{elu}.

Despite an apparent simplicity, neural networks are extremely expressive:  by using at least one layer of nonlinear activation functions, a sufficiently large network can represent \emph{any} function within a given bounded error~\cite{neuralNetsBook}. 

Instead of directly connecting a huge set of neurons to the input, it is common to organize them in groups of stacked layers that transform the outputs of the previous layer and feed it to the next layer. This enforces the networks to learn hierarchical features, performing low-level reasoning in the first layers 
(such as edge detection)
and higher-level tasks in the last layers
(e.g.~, assembling object parts). 
For this reason, the first and last layers are often referred to as lower and upper layers, respectively.

In an image categorization problem, the input of the network is an image (or a set of features derived from an image), and the goal is to predict the correct category of the entire image. We can view the pixelwise semantic labeling problem as taking an image patch and categorizing its central pixel. Finding the optimal neural network classifier reduces to finding the weights and biases that minimize a loss $L$ between the predicted labels and the target labels in a training set. 
Let $\mathcal{L}$ be the set
of possible semantic classes;
labels are typically encoded as a vector of length $|\mathcal{L}|$ with value `1' at the position of the correct label and `0' elsewhere. The network contains thus as many output neurons as possible labels. A softmax normalization is performed on top of the last layer to guarantee that the output is a probability distribution, i.e. the label values are between zero and one and sum to one. The multi-label problem is then seen as a regression on the desired output label vectors.

The loss function $L$ quantifies the misclassification by comparing the target label vectors $\mathbf{y}^{(i)}$ and the predicted label vectors $\mathbf{\hat{y}}^{(i)}$, for $n$ training samples $i=1\dots n$. In this work we use the common cross-entropy loss, defined as:
\begin{equation}
L = -\frac{1}{n}\sum\limits_{i=1}^{n} \sum\limits_{k=1}^{|\mathcal{L}|}y_k^{(i)} \log{\hat{y}_k^{(i)}}.
\label{eq:crossentropy}
\end{equation}
Training neural networks by optimizing this criterion converges faster
(compared with, for instance, the Euclidean distance between $\mathbf{y}$ and $\mathbf{\hat{y}}$). In addition, it is numerically stable when coupled with softmax normalization \cite{neuralNetsBook}. 

Once the loss function is defined, the parameters (weights and biases) that minimize the loss are found via gradient descent, by computing the derivative $\frac{\partial L}{\partial w_i}$ of the loss function with respect to every parameter $w_i$, and updating the parameters with a learning rate $\lambda$ as follows:
\begin{equation}
w_i \leftarrow w_i - \lambda \frac{\partial L}{\partial w_i}.
\end{equation}
The derivatives $\frac{\partial L}{\partial w_i}$ are obtained by \emph{backpropagation}, which consists in explicitly computing the derivatives of the loss with respect to the last layer's parameters and using the chain rule to recursively compute 
the derivatives of each layer's outputs with respect to its weights and inputs (i.e.~the lower layer's outputs).
In practice, 
instead of averaging over the full dataset,
the loss (\ref{eq:crossentropy}) 
is estimated
from a random
small subset of the training set, referred to as a mini-batch.
This learning technique is named \emph{stochastic gradient descent}.

\subsection{Convolutional Layers}
\label{s:convnets:convnets}

Convolutional neural networks (CNNs) \cite{lecunNets} contain so-called convolutional layers, a specific type of layer that imposes a number of restrictions compared to a more general fully-connected layer (discussed below). These restrictions (e.g., local connectivity) have been introduced for image categorization networks because they make sense in that particular context.


In CNNs, each neuron is associated to a spatial location $(i,j)$ in the input image. The output $a_{ij}$ associated with location $(i,j)$ in a convolutional layer is computed as:
\begin{equation}
a_{ij} = \sigma ( (\mathbf{W}\ast \mathbf{X})_{ij} + b),
\end{equation}
where $\mathbf{W}$ denotes a kernel with learned weights, $\mathbf{X}$ the input to the layer and `$\ast$' the convolution operation. Notice that this is a special case of the neuron in Eq.~\ref{eq:neuron} with the following constraints:
\begin{itemize}
\item The connections only extend to a limited spatial neighborhood determined by the kernel size;
\item The same filter is applied to each location, guaranteeing translation invariance.
\end{itemize}
Multiple convolution kernels are usually learned in every layer, interpreted as a set of spatial feature detectors. The responses to every learned filter are thus referred to as \emph{feature maps}. Note that the convolution kernels are actually three-dimensional:
in addition to their spatial extent (2D), 
they span along all the feature maps in the previous layer (or eventually through all the bands in the input image). As this third dimension can be inferred from the previous layer it is rarely mentioned in the architecture descriptions.

Compared to the fully connected layer, in which every neuron is connected to all outputs of the previous layer, a convolutional layer highly reduces the number of parameters by enforcing the aforementioned constraints. 
This results in 
an easier optimization problem,
without losing much generality. This opened the
door
to using the image itself as an input without any feature design and selection process,
as 
CNNs discover the relevant spatial features to conduct classification.

\subsection{Increasing the Receptive Field}

In CNNs, the \emph{receptive field} denotes the spatial extent of the input image connected to a certain neuron, possibly indirectly through other neurons in previous layers:
it is the set of pixels on which this neuron depends.
In other words, 
it
 quantifies how far a neuron can ``see'' in the image. In most applications, a large amount of spatial context must be taken into account in order to successfully label the images. For example, to deduce that a certain pixel belongs to a rooftop, it might not be enough to just consider its individual spectrum: 
we
might need to observe a large patch around this pixel, taking into account geometry and structure of the objects, to infer its correct class. 

Neural networks for image analysis should thus be designed to accumulate,
through their layers, a large enough receptive field. 
While a straightforward way to do it is to use large convolution kernels, this is not a common practice mostly due to its computational complexity. Besides, this would aim at learning large  filters all at once, 
with millions of parameters.
However, it is preferable to learn a hierarchy of small filters instead, reducing the number of parameters while remaining expressive, and thus making the optimization problem easier.

The most common approach to 
reduce the number of parameters for a given receptive field size
is to
  downsample
the feature maps throughout the network. This is commonly achieved progressively through interleaving downsampling layers with convolutional layers. This way, the resolution of the feature maps gets lower and lower as we traverse the layers from input to output. For example, 
neurons after a chain of two $3\times 3$ convolutions in successive layers would normally have a receptive field of $5\times 5$ pixels, which extends to $12\times 12$ pixels with an accumulated downsampling of factor 4.

To downsample the feature maps, the most popular approach is to use the so-called \emph{max pooling} layer~\cite{poolingTheoretical}. A max pooling layer takes a group of neighbors in the feature map and condenses them into a single output by computing the maximum of all incoming activations in the window. The pooling windows in general do not overlap, hence the output map is downsampled (see Fig.~\ref{f:fullyconvnet}). For instance, if pooling is performed in a $2\times 2$ window, the feature map is reduced to half of its resolution. 

Computing the maximum value is inspired by the idea of detecting objects from their parts. For example, in a face detector it is important to identify the constituents of a face, such as \emph{hair} or \emph{nose}, while the exact locations of these components should not be such a determinant factor. The max pooling layer conveys then to which extent there is evidence of the \emph{existence} of a feature in a vicinity. 
Other less popular forms of downsampling include average pooling and applying convolutions with a \emph{stride}, i.e., ``skipping'' some of them (e.g., applying every other convolution).

Pooling operations (and downsampling in general) hard-code robustness to spatial deformations, an attribute that boosted the success of CNNs for image categorization. However, spatial precision is lost when downsampling. The increased receptive field (and thus recognition capability) comes at the price of losing localization capability. This well-reported trade-off~\cite{fcn,deeplab} is a major concern for dense 
 labeling.

We could still imagine a downsampling network that preserves localization: it would learn features of the type ``a corner at the center of the receptive field'', ``a corner one pixel left of the center of the receptive field'', ``a corner two pixels left of the center of the receptive field'', 
and so on,
multiplying the number of features to be learned. 
This would however discredit the use of downsampling to gain robustness to spatial variation in the first place. 
The recognition/localization trade-off must thus be properly addressed to design a high-resolution semantic labeling network.

\subsection{Fully Convolutional Networks (FCNs)}
\label{s:convnets:fcns}

\begin{figure}
\centering
\includegraphics[scale=1.6]{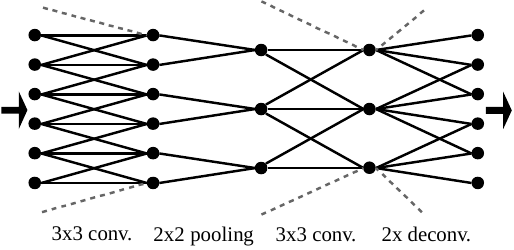}
\caption{Lateral view of a fully convolutional network (dashed lines indicate inputs that have been padded in conv. layers and cropped in the deconv. layer to preserve spatial dimensions).}
\vspace{-5pt}
\label{f:fullyconvnet}
\end{figure}

Image categorization networks are typically written as a series of interleaved convolution and pooling layers that extract spatial features, followed by a few fully connected layers that compute the final classification values. The dense semantic labeling problem can be seen as classifying the central pixel of an image patch, the size of the input patch being the receptive field used to classify it. To label the whole image the prediction must thus be performed on many overlapping image patches, 
which requires a huge amount of redundant operations.

Fully convolutional networks (FCNs) \cite{fcn} are especially relevant to semantic labeling. They contain only convolutional layers, i.e., no fully connected layers. 
Therefore, they can be applied to images with various sizes: inputting a larger image patch produces a larger output, the convolutions being performed on more locations. In contrast, networks with any fully connected layer require a fixed image size, because of the fixed input size of such layers.
Using fully convolutional networks also 
removes any redundancy when computing classification maps on large inputs,
as they are applied only once.

The first obvious advantage of FCNs is a reduced computational complexity. Moreover, we can efficiently train on input patches that are larger than the receptive fields, and in turn produce larger classified patches, with more than a single pixel. While the elements inside a contiguous patch are highly correlated, the use of moderately larger patch sizes has been reported to be beneficial~\cite{sherrah,kampf}. Furthermore, the patch size at training time is decoupled from the one at test time. For example, we could use use small patches to train the network in order to have a highly variable input in every mini-batch, but later conduct predictions on the largest patch size that fits in the GPU. Let us finally remark that  a traditional classification network (with fully connected layers) can be in fact easily rewritten as a fully convolutional network~\cite{fcn}. 

When an FCN has downsampling layers, the output contains 
fewer
elements than the input, since 
the
resolution has been decreased. This gave birth to the so-called deconvolutional (or upconvolutional) layer, which upsamples a feature map by interpolating neighboring elements (as the last layer in Fig.~\ref{f:fullyconvnet}). Instead of determining a priori the type of interpolation, e.g., bilinear, the operation is parametrized by a kernel that can also be learned. Deconvolutional layers are typically used to perform a naive interpolation at the very end of the pipeline, on the output classification maps.
In the next section we study more advanced ways of providing high-resolution outputs.

\section{Analysis of High-Resolution Labeling CNNs}
\label{s:analysis}

Fully convolutional networks (FCNs), as described in Section~\ref{s:convnets:fcns}, have become the standard in semantic labeling.
 Nevertheless, the open question is how to conduct fine predictions that provide detailed high-resolution outputs, while still taking large amounts of context into account and without exploding the number of trainable parameters. Simply adding a deconvolutional layer to upsample the output on top of a network provides dense outputs but imprecise labeling results, because the upsampling is performed in a naive way from the coarse classification. 
This is dissatisfying in many applications, such as high-resolution aerial image labeling, where the goal is to precisely identify and outline tiny objects such as cars.

\begin{figure}
\begin{subfigure}{0.48\linewidth}
\centering
\includegraphics[scale=0.055]{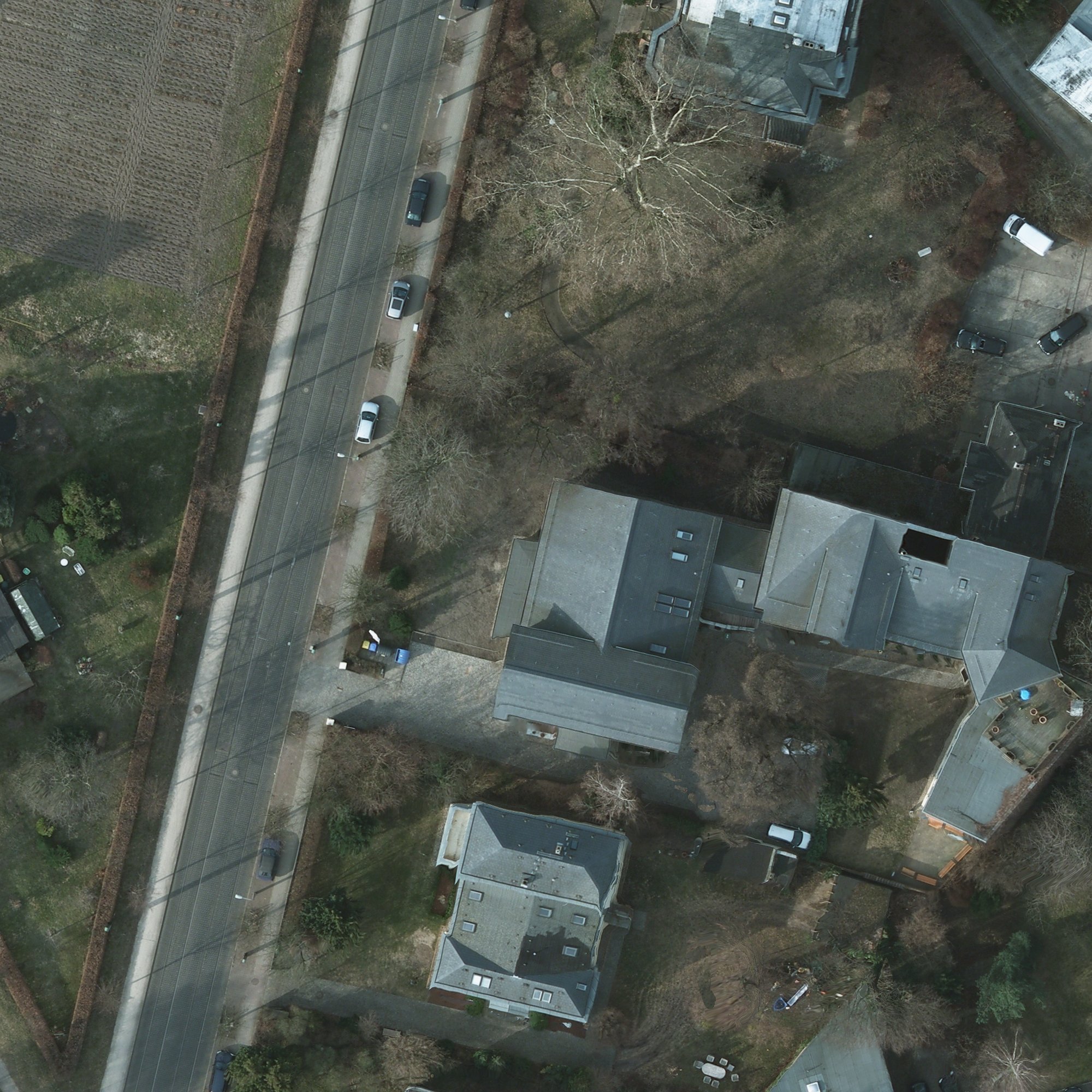}
\caption{}
\end{subfigure}
\begin{subfigure}{0.48\linewidth}
\centering
\includegraphics[scale=0.055]{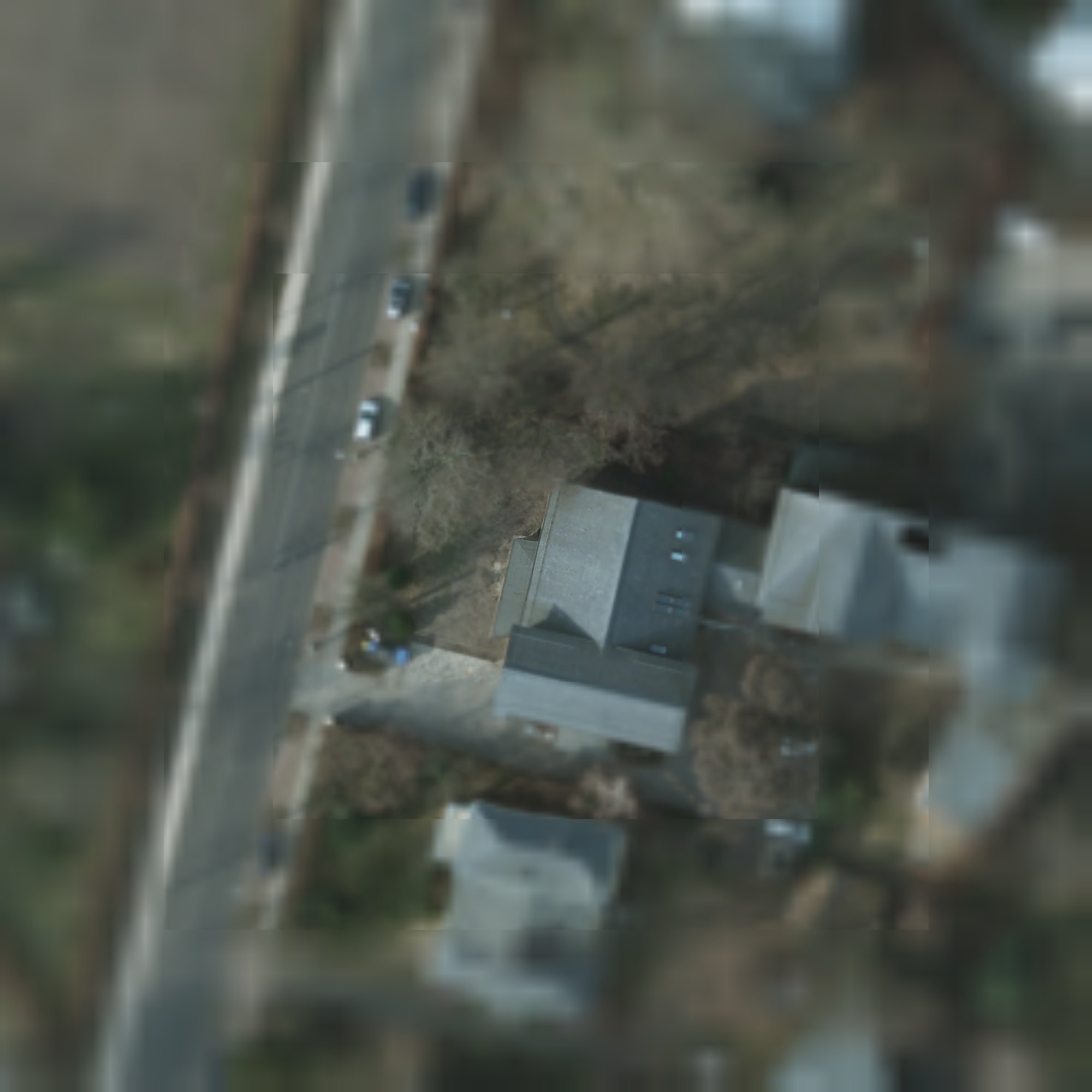}
\caption{}
\end{subfigure}
\centering
\caption{To classify the central gray pixel of this patch (and not to confuse it, e.g., with an asphalt road), we need to take into account a spatial context (a). However, we do not need a high resolution everywhere in the patch. It can be lower as we go away from the central pixel and still identify the class (b).}
\label{f:fullnotfullres}
\vspace{-5pt}
\end{figure}

We now describe what we consider to be the elementary principle from which to derive efficient semantic labeling architectures.
Let us then first observe that while our goal is to take large amounts of context into account, we do not need this context at the same spatial resolution everywhere. For example, let us suppose we want to classify the central pixel of the patch in Fig.~\ref{f:fullnotfullres}a. Such a gray pixel,
taken out of context,
could be easily confused with an asphalt road. Considering the whole patch at once helps to infer that the pixel belongs indeed to a gray rooftop. 
However, two significant issues arise if we take a full-resolution large patch for context: a) it requires many computational resources that are actually not needed for an effective labeling, and b) it does not provide robustness to spatial variation (we might actually not care about the exact location of certain features). 
Conducting predictions from low-resolution patches instead is not a solution as it produces inaccurate coarse classification maps.
Nevertheless, it is actually not necessary to observe all surrounding pixels at full resolution: the farther we go from the pixel we want to label, the lower the requirement to know the exact location of the objects. For example, in the patch of Fig.~\ref{f:fullnotfullres}b it is still possible to classify the central pixel, despite the outer pixels being blurry. 
Therefore, we argue that a combination of reasoning at different resolutions is necessary to conduct fine labeling, if we wish to take a large context into account in an efficient manner.


In the following, we analyze the main families of high-resolution semantic labeling networks that have been proposed in the past two years. For each of them we discuss the following aspects:
\begin{itemize}
\item How a solution to the fine-grained labeling problem is provided;
\item Where this solution stands with respect to the principle of Fig.~\ref{f:fullnotfullres};
\item General advantages and disadvantages, and computational efficiency.
\end{itemize}


\subsection{Dilation Networks}

\begin{figure}
\centering
\includegraphics[scale=2]{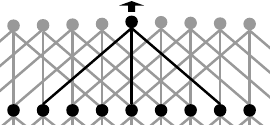}
\caption{A dilated convolution (i.e., on non-adjacent inputs) with a dilation factor $S=4$.}
\label{f:dilated}
\vspace{-9pt}
\end{figure}

Dilation networks are based on the shift-and-stitch approach or \emph{\`a trous} algorithm~\cite{fcn}. This  consists in conducting a prediction at different offsets to produce multiple low-resolution outputs, which are then interleaved to compose the final high-resolution result. For example, if the downsampling factor of a network is $S$, one should obtain $S^2$ classification maps by shifting the input horizontally and vertically. Such an interleaving can also be implemented directly in the architecture, by using ``dilated'' operations~\cite{dilatedConv}, i.e., performing them on non-contiguous elements of the previous feature maps. This principle is illustrated in Fig.~\ref{f:dilated}. 

Dilations have been used with two purposes:
\begin{enumerate}
\item as an alternative to upsampling for generating full-resolution outputs~\cite{mammography,fcn},
\item as a means to increase the receptive field~\cite{deeplab,dilatedConv}, by enlarging the area covered by a convolution kernel without increasing the number of trainable parameters.
\end{enumerate}

Regarding the first point, we must mention that there is no theoretical improvement compared to an FCN with naive upsampling, because the presence of pooling layers still reduces spatial precision. 
 Executing the prediction multiple times at small offsets still keeps predictions spatially imprecise.

Regarding the second point, we must remark that while dilated convolutions increase the receptive field, this does not introduce robustness to spatial variation per se. For example, a network with only dilated convolution layers would have a large receptive field but would only be able to learn filters of the type ``a building in the center, with a car \emph{exactly} five pixels to the left''. This robustness would have  to be thus learned, hopefully, by using a larger number of filters. 

The use of an interleaved architecture at training time, implemented with dilations,  has been however reported to be beneficial. In the context of aerial image labeling, Sherrah~\cite{sherrah} recently showed that it outperformed its FCN/upsampling counterpart\footnote{While such architecture is named a ``no-downsampling'' network in~\cite{sherrah}, a more appropriate name would probably be ``no-upsampling'', because there is indeed downsampling due to the max pooling layers.}.
 The major improvement compared to the FCN/upsampling network was measured in the labeling capabilities of the \emph{car} class, which is a minority class with tiny objects, difficult to recognize~\cite{volpi}.
  While the dilation strategy is not substantially different from an architectural point of view compared to naive upsampling, some advantages in training might explain the better results: In the upsampling case the network is encouraged to provide a coarse classification that, once upsampled, is close to the ground truth.  In the dilation network, on the contrary, the interleaved outputs are directly compared to individual pixels in the ground truth, one by one. The latter seems to better avoid suboptimal solutions that absorb minority classes or tiny objects.

The computational time and memory required by dilation networks are significant, to the point that using GPUs might become impractical even with moderately large architectures.
This is because the whole network rationale is applied at many contiguous locations.

Overall, while dilation networks have been reported to exhibit certain advantages, they are computationally demanding and do not particularly address the principle of Fig.~\ref{f:fullnotfullres}.

\subsection{Deconvolution Networks (unpooling)}
\label{s:analysis:deconv}

\begin{figure}
\centering
\includegraphics[scale=1.6]{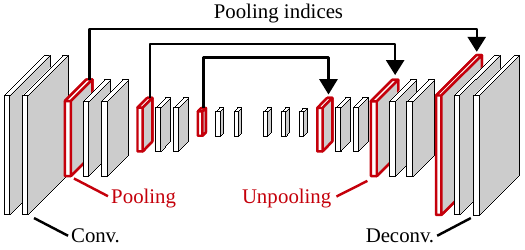}
\caption{Deconvolution network. The CNN is ``mirrored'' to learn the deconvolution.}
\label{f:deconvnet}
\vspace{-5pt}
\end{figure}

\begin{figure}
\centering
\includegraphics[scale=1.65]{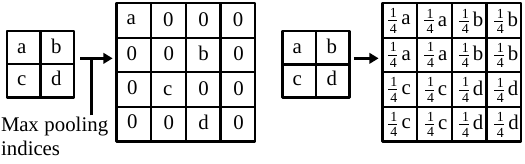}
\caption{Max (left) and average (right) unpooling.}
\label{f:unpooling}
\end{figure}

Instead of naively upsampling the classification score maps with one deconvolutional layer, a more advanced approach is to attach a multi-layer network to learn a complex upsampling function. This idea was simultaneously presented by different research groups \cite{unpooling,segnet1} and later extended to different problems (e.g.,~\cite{contourdetect}). The most hassle-free way to do this is to simply ``reflect'' an existent FCN, with the same number of layers and kernel sizes, to perform the upsampling.
The convolutional layers are reflected as deconvolutional layers, and the pooling layers as \emph{unpooling} layers (see Fig.~\ref{f:deconvnet}). 
While pooling condenses several activations into one representative value (typically, the maximum activation), unpooling layers must reconstruct 
the original size of activations. 
In the case of max unpooling, the location of the maximal activation is recalled from the corresponding pooling layer, and is used to place the activation back into its original pooled location. The other elements in the unpooling window are set to zero, leading to sparse feature maps, as illustrated in Fig.~\ref{f:unpooling}. Unpooling was first introduced as part of a framework to analyze and visualize CNN features~\cite{visualizingcnn}. The arrows in Fig.~\ref{f:deconvnet} represent the communication of the pooling indices from the pooling layer to the unpooling layer. In the case of average pooling, the corresponding unpooling layer simply outputs at every location the input activation value divided by the number of elements in the target unpooling window(see Fig.~\ref{f:unpooling}). In this case, there is no need to transmit a location from pooling to unpooling.

This concept can be thought of as an ``encoder--decoder'',
where the middle layer is seen as a common representation to images and classification maps, while the ``encoder'' and ``decoder'' ensure the translation between this representation and the two modalities.
When converting an FCN to a deconvolution network, the final classification layer of the FCN is usually dropped 
before
reflecting the architecture. This way the interface between the encoder and the decoder is a rich representation with multiple features.
The first layer of the encoder takes as input as many channels there are in the input image, and usually the last layer of the decoder produces as many feature maps as classes required. In \cite{segnet1,segnet2}, alternatively, the network outputs a larger set of features that are then classified with additional layers. 

While pooling is used to add robustness to spatial deformation, the fact of ``remembering'' the location of the max activation helps to precisely locate objects in the deconvolution steps. For example, the exact location of a road might be irrelevant to do any higher-level reasoning later on, but once the network decides to label the road as a semantic object we need to recover the location information to outline it with high precision. This illustrates how deconvolution networks balance the localization/recognition trade-off.

Note however that if one happens \emph{not} to choose max pooling for downsampling, then the unpooling scheme is not able to recover \textit{per se} the lost spatial resolution. There is no memory about the location of the higher resolution feature. Even though max pooling is very common, it has been shown that average or other types of pooling might be more effective in certain applications~\cite{poolingTheoretical}. In fact, recent results~\cite{allconv} suggested that  max pooling can be emulated with a strided convolution and achieve similar performance. The deconvolution network idea is however leveraged when max pooling is the downsampling mechanism used.

This certainly does not mean that a deconvolutional network is incapable of learning without max pooling layers. Convolution/deconvolution architectures without max pooling have been successfully used in different domains \cite{volpi,sketchsimpl}. For example, a recent submission to the ISPRS Semantic Labeling Challenge~\cite{volpi} is such type of network.
 The recognition/localization trade-off is not really alleviated in this case: the encoder should encode features of the type ``an object boundary 5 (or 7, 10...) pixels away to the left'', so that the decoder can really leverage that information and reconstruct a high-resolution classification map. 

The depth of deconvolution networks is significantly larger, roughly twice the one of the associated FCN. This often implies a slower and more difficult optimization, due to the increase in trainable parameters introduced by deconvolutional layers. 
While the decoding part of the network can be simplified~\cite{realtimenet}, 
this adds arbitrariness to the design.

To conclude, the deconvolution scheme does address the recognition/localization trade-off, but only in
  the case where max pooling is used for downsampling. The increased network depth can be a concern for an effective training.

\subsection{Skip Networks}
\label{s:analysis:skip}

In the original paper about fully convolutional networks, Long et al.~\cite{fcn} proposed the so-called ``skip'' architecture to  generate high-resolution classification outputs.
The idea is to 
build  
the final classification map by combining multiple classification maps, obtained from intermediate features of the network at different resolutions (and not just the last one). 

The last layer of an FCN outputs as many feature maps as classes, which are interpreted as score or ``heat'' maps for every class. Intermediate layers, however, tend to have many more features than the number of classes. Therefore, skip networks add extra layers that convert the arbitrarily large number of features of intermediate layers into  the desired number of heat maps. This approach allows us to extract multiple score maps for each class from a single network, at different resolutions. The lower-level score maps are fine but have a small receptive field, while the higher-level ones can see farther but with less detail. As a result, we have a pool of score maps.

\begin{figure}
\centering
\includegraphics[scale=1.5]{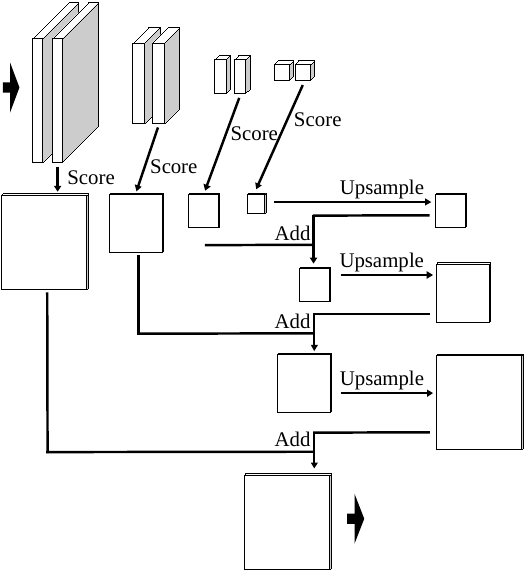}
\caption{Skip network: multiple classification scores are obtained from intermediate CNN features at different resolutions, and are combined by element-wise adding and upsampling.}
\vspace{-5pt}
\label{f:skip}
\end{figure}

The score maps are then combined pairwise, from the lower scales to the higher scales. At every step, the lower-resolution score maps are upsampled to match the higher-resolution ones. They are then added elementwise.
This is repeated until all intermediate maps are processed. 
 The overall combination of resolutions forms a directed acyclic graph, with links that ``skip'' ahead from lower layers to higher ones. A skip network is illustrated in Fig.~\ref{f:skip}.

Skip networks address the trade-off between localization and precision quite explicitly: the information at different resolutions is extracted and combined. The original paper introduces 
this methodology as ``combining what and where''. This approach is closer to the principle described in Fig.~\ref{f:fullnotfullres} than the previous approaches reviewed above. The skip network mixes observations at different resolutions, without unnecessarily increasing the depth or width of the architecture (as in deconvolution and dilation networks respectively) and it does not impose a particular type of downsampling (as in deconvolution networks).

While the idea of extracting different resolutions is certainly very relevant, the skip model seems to be inflexible and arbitrary in how to combine them. First of all, it combines  classification verdicts, instead of a rich set of features, coming from each of the resolutions.   For example, it combines how a layer evaluates that an object is a building by using low-level information, with how another layer evaluates whether the same object is a building by using higher-level information. Let us recall that we use deep multi-layer schemes with downsampling because we actually consider that certain objects can only be detected at the upper layers of the network, when a large amount of context has been taken into account and at a high level of abstraction. It seems thus contradictory to try to refine the boundaries of an object detected at a high level, by using a classification conducted at a lower level, where the object might not be detected at all. 
Moreover, the element-wise addition restricts the combination of resolutions to be simply a linear combination. The skip links to combine resolutions are in fact parameterless (besides the addition of the scoring layers). We could certainly imagine classes that require a more complex nonlinear combination of high- and low-level information to be effectively classified.

It is worth noting that the creation of the intermediate score maps has also been referred to as a dimensionality reduction step~\cite{segnet2}. 
 It is however not
by chance that the amount of reduced features coincides with the amount of classes:
even though it is technically a dimensionality reduction, its spirit is to create a partial classification, not just to reduce the number of features. 
This is confirmed by the name of these layers in  the public implementation of \cite{fcn}~: ``score'' layers.
Moreover, if this operation were indeed intended to be just a reduction of dimensionality, we could imagine outputting different amounts of feature maps from different resolutions. However, in that case there would be no way of adding them element by element as suggested.

To conclude, the skip network architecture provides an efficient solution to address the localization/recognition trade-off, yet this could be done in a more flexible way that allows a more complex combination of the features.

\section{Learning to Combine Resolutions}
\label{s:mlp}

In this section we propose an alternative scheme for high-resolution labeling, derived as a natural consequence of our observations about the other families of methods. In particular, this architecture leverages the benefits of the skip network described in Section~\ref{s:analysis:skip} while addressing its potential limitations.  

Taking multiple intermediate features at different resolutions and combining them seems to be a sensible approach to specifically address the localization/recognition trade-off, as done with skip networks. In such a scheme, the high-resolution features have a small receptive field, while the low-resolution ones have a wider receptive field. Combining them constitutes indeed an efficient use of resources, since we do not actually \emph{need} the high-resolution filters to have a wide receptive field, following the principle of Fig.~\ref{f:fullnotfullres}.

\begin{figure}
\centering
\includegraphics[scale=1.5]{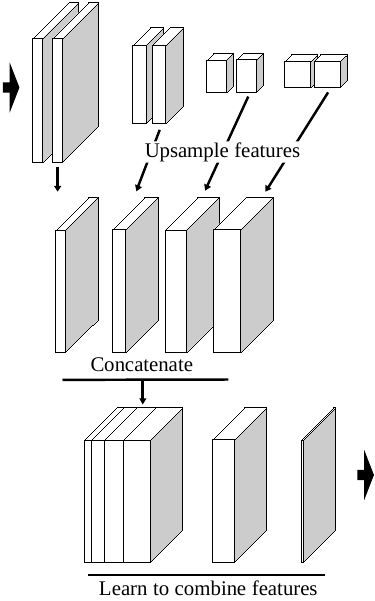}
\caption{MLP network: intermediate CNN features are concatenated, to create a pool of features. Another network learns how to combine them to produce the final classification.}
\label{f:mlp}
\end{figure}

The skip network combines \emph{predictions} derived from the different resolutions, i.e., score maps for each of the classes. For example, we try to refine the ``blobby'' building outputted by the coarse classifier, via a higher-resolution classification. However, it is unclear how effectively the higher-resolution classifier detects the building in question, considering its reduced receptive field and shallow reasoning.

We thus argue that the most appropriate way of performing fine semantic labeling is to combine features, not classification maps. For example, to refine the boundaries of a coarse building, we would use high-resolution edge detectors (and not high-resolution building detectors).

In our proposed scheme,  intermediate features are extracted from the network and treated equally, creating a pool of features that emanate from different resolutions. A neural network then learns  how to combine these features to give the final classification verdict. This adds flexibility to learn more complex relations between the different resolutions and generalizes the element-wise addition of the skip architecture. 

The overall process is depicted in Fig.~\ref{f:mlp}. First, a subset of intermediate features are extracted from the network. These are naively upsampled to match the resolution of the higher-resolution features. They are then concatenated to create the pool of features. Notice that while the spatial dimensions of the feature maps are all the same, they originally come from different resolutions. This way, the variation of the feature responses across space will be smoother in certain maps while sharper in others. Note that while it is practical to store in memory the upsampled responses, this is not intrinsically necessary. For example, we could imagine a system that answers to a high-resolution query by outputting the nearest neighbor in the coarser map or by interpolating neighboring values on the fly.

From the pool of features, a neural network predicts the final classification map (we could certainly use other classifiers, but this lets us train the system end to end).  We assume that all the spatial reasoning has been conveyed in the features computed by the initial CNN. This is why  we operate on a pixel-by-pixel basis to combine the features.
 Any need to look at 
neighbors should be expressed in the spatial filters of the CNN. This way we conceptually and architecturally separate the extraction of spatial features from their combination. 

We can think of the multi-layer perceptron (MLP) with one hidden layer and a non-linear activation function as a minimal system to learn how to combine the pool of features. Such MLPs can learn to approximate any function and, since we do not have any particular constraints, it seems an appropriate choice. In practice, this is implemented as a succession of convolutional layers with $1 \times 1$ kernels, since we want the same MLP to be applied at every location. By introducing the MLP and executing it at a fine resolution, we must expect an overhead in processing time compared to the skip network.

The proposed technique is intended to learn how to combine information at different resolutions, not how to upsample a low-resolution classification. An example of the type of relation we are able to convey in this scheme is as follows:  ``label a pixel as \emph{building} if it is red and belongs to a larger red rectangular structure, which is surrounded by areas of green vegetation and near a road''. 

Finally, let us discuss the CNN from which features are extracted (the topmost part of Fig.~\ref{f:mlp}). The different features are extracted from intermediate layers of a single CNN. This assumes that the higher-level features can be derived from the lower-level ones. It is basically a part-based model~\cite{felzenszwalb2010object}, where we consider that an object can be detected by its parts, and we are using those same parts as the higher-resolution features inputted to the MLP. This seems to be a sensible assumption, yet we must mention that we could eventually think of separate networks to detect features at different resolutions instead of extracting intermediate representations of a single network (as, e.g., in~\cite{farabetSceneLabeling}). While we adopt  the model of Fig.~\ref{f:mlp} in this work, the alternative could be also considered. It would be certainly interesting to study to which extent it is redundant to learn the features in separate networks and, conversely, how results could be eventually improved by doing it.



\section{Experiments}
\label{s:experiments}

\subsection{Datasets and Evaluation Metrics}

We evaluate the aforementioned architectures on two benchmarks of aerial image labeling: Vaihingen and Potsdam, provided by Commission III of the ISPRS~\cite{2dSemLabeling}. The Vaihingen dataset is composed of 33 image tiles (of average size $2494\times 2064$), out of which 16 are fully annotated with class labels. The spatial resolution is 9 cm. Near infrared (NIR), red (R) and green~(G) bands are provided, as well as a digital surface model (DSM), normalized and distributed by~\cite{isprs_gerke}. We select 5 images for validation (IDs: 11, 15, 28, 30, 34) and the remaining 11 images for training, following~\cite{sherrah,volpi,isprs_effective}. 

Potsdam dataset consists of 38 tiles of size $6000\times 6000$ at a spatial resolution of 5 cm, out of which 24 are annotated. It provides an additional blue channel and the normalized DSM. We select the same 7 validation tiles as in~\cite{sherrah} (IDs: 2\_11, 2\_12, 4\_10, 5\_11, 6\_7, 7\_8 7\_10) and the remaining 17 tiles for training. Both datasets are labeled into the following six classes: \emph{impervious surface}, \emph{building}, \emph{low vegetation}, \emph{tree}, \emph{car} and \emph{clutter/background}.

In order to account for labeling mistakes, another version of the ground truth with eroded boundaries is provided, on which accuracy is measured. To evaluate the overall performance, overall accuracy is used, i.e., the percentage of correctly classified pixels. To evaluate class-specific performance, the F1-score is used, computed as the harmonic mean between precision and recall~\cite{acc}. We also include the mean F1 measure among classes, since overall accuracy tends to be less sensitive to minority classes in imbalanced datasets.

\subsection{Network Architectures}
\label{s:exp:netarchi}

\begin{table}
\caption{Architecture of our base FCN.}
\centering
\begin{tabular}{ccccc}
Layer & Filter size & Number of filters & Stride & Padding \\ 
\hline 
Conv-1\_1 & 5 & 32 & 2 & 2 \\ 
Conv-1\_2 & 3 & 32 & 1 & 1 \\ 
Pool-1 & 2 &  & 2 &  \\ 
Conv-2\_1 & 3 & 64 & 1 & 1 \\ 
Conv-2\_2 & 3 & 64 & 1 & 1 \\ 
Pool-2 & 2 &  & 2 &  \\ 
 Conv-3\_1 & 3 & 96 & 1 & 1 \\ 
Conv-3\_2 & 3 & 96 & 1 & 1 \\ 
Pool-3 & 2 &  & 2 &  \\ 
 Conv-4\_1 & 3 & 128 & 1 & 1 \\ 
Conv-4\_2 & 3 & 128 & 1 & 1 \\ 
Pool\_4 & 2 &  & 2 &  \\ 
Conv-Score & 1 & 5 & 1 &  \\ 
\hline 
\end{tabular} 
\label{t:basefcn}
\end{table}

To conduct our experiments we depart from a base fully convolutional network (FCN) and derive other architectures from it. Table~\ref{t:basefcn} summarizes our base FCN for the Vaihingen dataset. The architecture is borrowed from~\cite{kampf}, except for the fact that we increased the size of the filters from 3 to 5 in the first layer, since it is a common practice to use larger filters if there is a stride. 
Every convolutional layer (except the last one) is followed by a batch normalization layer~\cite{batchnorm} and a ReLU activation. 
We did not optimize the architecture of the base FCN.

The total downsampling factor is 16, out of which 8 is the result of the max pooling layers and 2 of the stride in the first layer. The conversion of the last set of features to classification maps (the ``score'' layer) is performed by a $1\times 1$ convolution. To produce a dense pixel labeling we must add a deconvolutional layer to upsample the predictions by a factor of 16, thus bringing them back to the original resolution.

To implement a \emph{skip} network, we extract the features of layers \emph{Conv-*\_2}, i.e., produced by the last convolution in each resolution and before max pooling. Additional scoring layers are added to 
produce 
classification maps from the intermediate features. The resulting score maps are then combined as explained in Section~\ref{s:analysis:skip}.
Our \emph{MLP} network was implemented by extracting the same set of features. As no intermediate scores are needed, we remove layer `Conv-Score' from the base FCN.  The features are combined as explained in Section~\ref{s:mlp}. The added multi-layer perceptron contains one hidden layer with 1024 neurons.

We also created a deconvolution network that exactly reflects the base FCN (as in \cite{unpooling}). This is straightforward, with deconvolutional and unpooling layers associated to every convolutional and pooling layer. The only particularity is that the last layer outputs as many maps as required classes and not as input channels. We here call it \emph{unpooling} network, to differentiate it in the experiments from another method that uses a stack of deconvolutions but without unpooling~\cite{volpi}, which we simply refer to as \emph{deconvolution} network. To cover the last family of architectures of Sec.~\ref{s:analysis}, the \emph{dilation} network, we incorporate the results recently presented by Sherrah~\cite{sherrah}.

In both datasets we use the same four input channels: DSM, NIR, R and G. Notice that we simply add the DSM as an extra band. While for Potsdam dataset the blue channel is also available, we here excluded it for simplicity. In the case of Vaihingen we predict five classes, ignoring the clutter class, due to the lack of training data for that class. In the case of Potsdam we predict all six classes.

Considering the difference in resolution in both datasets, in the case of Potsdam we downsample the input and linearly upsample the output by a factor of 2 (following~\cite{sherrah}). We use the same architecture as for Vaihingen (besides the different number of output classes) 
between the
downsampling and upsampling layers. This is 
intended 
to cover similar receptive field in terms of meters (and not pixels) for both datasets.

\subsection{Training}
\label{s:exp:training}

The networks are trained by stochastic gradient descent~\cite{neuralNetsBook}. In every iteration a group of  patches is fed to the network for backpropagation. We sample random patches from the images, performing random flips (vertically, horizontally or both) and transpositions, augmenting the data 8 times. At every iteration we group five patches in the mini-batch, of size $256\times 256$ for Vaihingen dataset and $512\times 512$ for Potsdam (to roughly cover the same geographical area, considering the difference in resolution). In all cases, gradient descent is run with a momentum of 0.9, and an L2 penalty on the network's parameters of 0.0005. 
Weights are initialized following~\cite{initialization} and, since we use batch normalization layers before ReLUs, there is no need to normalize the input channels.

We start from a base learning rate of 0.1 and anneal it with an exponential decay. The decay rate is set so that the learning rate is divided by ten every 10,000 iterations in the case of Vaihingen and every 20,000 iterations in Potsdam. We decrease the learning rate more slowly in the case of Potsdam because the total surface covered by the dataset is larger, thus we assume it must take longer to explore. Training is stopped after 45,000 iterations in the first dataset and 90,000 in the second one, when the error stagnates on the validation set.

To train the unpooling, skip and MLP networks we initialize the weights with the pretrained base FCN, and jointly retrain the entire architecture. We start this second training phase with a learning rate of 0.01, and stop after 30,000 and 65,000 iterations for Vaihingen and Potsdam datasets respectively. We verified that the initialization with the pretrained weights is indeed beneficial compared to training from scratch.

\subsection{Numerical Results}

In this section we first present how our base FCN network compares to its derived architectures: unpooling, skip and MLP. We then position MLP with respect to other results reported in the literature, including a dilation network, thus completing the evaluation over all four families of techniques. We finally discuss our submission to the ISPRS contest.

\begin{table*}
\caption{Numerical evaluation of architectures derived from our base FCN on the Vaihingen validation set.}
\centering
\begin{tabular}{cccccc|c|c}
 & Imp. surf. & Building & Low veg. & Tree & Car & Mean F1 & Overall acc. \\ 
\hline 
Base FCN  & 91.46 & 94.88 & 79.19 & 87.89 & 72.25 & 85.14 & 88.61  \\ 
Unpooling & 91.17 & 95.16 & 79.06 & 87.78 & 69.49 & 84.54 & 88.55 \\ 
Skip      & 91.66 & 95.02 & 79.13 & 88.11 & 77.96 & 86.38 & 88.80 \\ 
MLP       & \textbf{91.69} & \textbf{95.24} & \textbf{79.44} & \textbf{88.12} & \textbf{78.42} & \textbf{86.58} & \textbf{88.92} \\
\hline 
\end{tabular} 
\label{t:vaih:valid}
\end{table*}

\begin{table*}
\caption{Numerical evaluation of architectures derived from our base FCN on the Potsdam validation set.}
\centering
\begin{tabular}{ccccccc|c|c}
 & Imp. surf. & Building & Low veg. & Tree & Car & Clutter & Mean F1 & Overall acc. \\ 
\hline 
Base FCN  & 88.33 & 93.97 & 84.11 & 80.30 & 86.13 & 75.35 & 84.70 & 86.20  \\ 
Unpooling & 87.00 & 92.86 & 82.93 & 78.04 & 84.85 & 72.47 & 83.03 & 84.67 \\ 
Skip      & 89.27 & 94.21 & 84.73 & \textbf{81.23} & 93.47 & 75.18 & 86.35 & 86.89 \\ 
MLP       & \textbf{89.31} & \textbf{94.37} & \textbf{84.83} & 81.10 & \textbf{93.56} & \textbf{76.54} & \textbf{86.62} & \textbf{87.02} \\
\hline 
\end{tabular} 
\label{t:pots:valid}
\end{table*}

\paragraph{Comparison of a base FCN to its derived unpooling, skip and MLP networks}

The classification performances on the validation sets 
 are included in Tables~\ref{t:vaih:valid} and~\ref{t:pots:valid}, for Vaihingen and Potsdam datasets, respectively. The MLP network exhibits the best performance in almost every case. The skip network effectively enhances the results compared 
with  
the base network, yet it does not outperform MLP. Let us remark  that the unpooling strategy does not necessarily constitute an improvement to the base FCN. This might be a result of the increased training difficulty due to the depth of the network and the sparsity of the unpooled maps. We tried to modify the training scheme, yet we could not improve its performance. 
 
 Overall, the numerical results show that the injection of lower-resolution features significantly improves the classification accuracy. MLP is the most competitive method, boosting the performance by learning how to combine these features.


\paragraph{Comparison with other methods}
Tables~\ref{t:vaih:valid:others} and~\ref{t:pots:valid:others} (for Vaihingen and Potsdam datasets respectively) incorporate the numerical results reported by other authors using the same training and validation sets. Since not every method was applied to both datasets, the tables do not display exactly the same techniques. 
 The MLP approach also outperforms the dilation strategy, in both datasets, thus positioning it as the most competitive category among those presented in Sections~\ref{s:analysis},~\ref{s:mlp} (dilation, unpooling, skip, MLP).

In the case of Vaihingen dataset, we also report the results of the \emph{deconvolution} network~\cite{volpi}, commented in Sec.~\ref{s:analysis:deconv}, which performs upsampling by using a series of deconvolutional layers. Contrary to the \emph{unpooling} network, the decoder does not exactly reflect the encoder and no unpooling operations are used. Additionally, we include the performance of other methods recently presented in the literature: the CNN+RF approach~\cite{isprs_effective}, which combines a CNN with a random forest classifier; the CNN+RF+CRF approch, which adds CRF post-processing to CNN+RF; and Dilation+CRF~\cite{sherrah}, which adds CRF post-processing to the dilation network. As depicted in the table, the MLP approach outperforms these other methods too.

For Potsdam dataset, Table~\ref{t:pots:valid:others} reports the performance of two other methods, presented in~\cite{sherrah}. In both cases, a pretrained network based on VGG~\cite{vgg} is applied to the IR-R-G channels of the image, and another FCN is applied to the DSM, resulting in a huge hybrid architecture. An ordinary version (with upsampling at the end) and a \emph{dilation} version are considered (`VGG pretr.' and `VGG+Dilation' in Table~\ref{t:pots:valid:others}, respectively). In the latter version, the dilation strategy could only be applied partially as it is too memory intensive.
While MLP outperforms the non-pretrained simpler dilation network, the \emph{VGG+Dilation} variants exhibits the best overall performance (though not on all of the individual classes). This suggests that the VGG component might be adding a competitive edge, though the authors stated that this is not the case on the Vaihingen dataset. 

Overall, MLP provides better accuracies than most techniques presented in the literature, including dilation networks, ensemble approaches and CRF post-processing.

\paragraph{Submission to the ISPRS challenge}
We submitted the result of executing MLP on the Vaihingen test set to the ISPRS server (ID: `INR'), which can be accessed online~\cite{2dSemLabeling}. Our method scored second out of 29 methods, with an overall accuracy of 89.5. 
 Note that the MLP technique described in this paper is very simple compared to other methods in the leaderboard, yet it scored better than them. For example, an ensemble of two \emph{skip} CNNs was  pretrained on large natural image databases~\cite{isprs_ensemble}, with over 20 convolutional layers and separate paths for the image and the DSM.
 Despite being simpler, our MLP network outperforms it in the benchmark.

\begin{table*}

\begin{minipage}{0.67\linewidth}
\captionof{table}{Comparison of MLP with other methods on the Vaihingen validation set.}
\centering
\begin{tabular}{cccccc|c|c}
 & Imp. surf. & Build. & Low veg. & Tree & Car & F1 & Acc. \\ 
\hline 
CNN+RF \cite{isprs_effective}	  & 88.58 & 94.23 & 76.58 & 86.29 & 67.58 & 82.65 & 86.52  \\ 
CNN+RF+CRF \cite{isprs_effective}     & 89.10 & 94.30 & 77.36 & 86.25 & 71.91 & 83.78 & 86.89 \\ 
Deconvolution \cite{volpi}       &  &  &  &  &  & 83.58 & 87.83 \\ 
Dilation \cite{sherrah}     & 90.19 & 94.49 & 77.69 & 87.24 & 76.77 & 85.28 & 87.70 \\ 
Dilation + CRF \cite{sherrah}     & 90.41 & 94.73 & 78.25 & 87.25 & 75.57 & 85.24 & 87.90 \\ 
MLP           & \textbf{91.69} & \textbf{95.24} & \textbf{79.44} & \textbf{88.12} & \textbf{78.42} & \textbf{86.58} & \textbf{88.92} \\
\hline
\end{tabular} 
\label{t:vaih:valid:others}

\vspace{12pt}
\caption{Comparison of MLP with other methods on the Potsdam validation set.}
\begin{tabular}{ccccccc|c|c}
 & Imp. surf. & Build. & Low veg. & Tree & Car & Clutter & F1 & Acc. \\ 
\hline 
Dilation \cite{sherrah} & 86.52 & 90.78 & 83.01 & 78.41 & 90.42 & 68.67 & 82.94 & 84.14  \\ 
VGG pretr.  \cite{sherrah}    & 89.84 & 93.80 & 85.43 & 83.61 & 88.00 & 74.48 & 85.86 & 87.42 \\ 
VGG+Dilation \cite{sherrah}   & \textbf{89.95} & 93.73 & \textbf{85.91} & \textbf{83.86} & \textbf{94.31} & 74.62 & \textbf{87.06} & \textbf{87.69} \\ 
MLP       & 89.31 & \textbf{94.37} & 84.83 & 81.10 & 93.56 & \textbf{76.54} & 86.62 & 87.02 \\
\hline 
\end{tabular} 
\label{t:pots:valid:others}

\end{minipage}
\begin{minipage}{0.33\linewidth}
\captionof{table}{Execution times.}
\centering
\begin{tabular}{ccc|cc}
• & \multicolumn{2}{c}{Train [s]}  & \multicolumn{2}{c}{Test [s/ha]} \\ 
\cline{2-5} 
• & Vaih. & Pots. & Vaih. & Pots. \\ 
\hline 
Base FCN & 3.9  & 9.8  & 0.81  & 1.44 \\ 
Unpooling & 8.4  & 21.0  & 1.38  & 1.84  \\ 
Skip & 6.6  & 16.9  & 0.81  & 1.48  \\ 
MLP & 10.0  & 24.5  & 1.70  & 2.0  \\ 
Dilation* & 62 & 400  & 4.81  & 17.2  \\ 
\hline 
\end{tabular} 
\label{t:exectime}

\vspace{5pt}
\flushleft {  \footnotesize \hspace{5pt} *As reported in~\cite{sherrah}, using a faster machine (see details in Sec.~\ref{s:runningtimes}.) }

\end{minipage}
\end{table*}

\begin{figure*}
\centering

\begin{subfigure}{0.005\linewidth}

\end{subfigure}
\begin{subfigure}{0.16\linewidth}
\centering
Image
\end{subfigure}
\begin{subfigure}{0.16\linewidth}
\centering
Ground truth
\end{subfigure}
\begin{subfigure}{0.16\linewidth}
\centering
Base FCN
\end{subfigure}
\begin{subfigure}{0.16\linewidth}
\centering
Unpooling
\end{subfigure}
\begin{subfigure}{0.16\linewidth}
\centering
Skip\end{subfigure}
\begin{subfigure}{0.16\linewidth}
\centering
MLP
\end{subfigure}

\begin{subfigure}{0.005\linewidth}
1
\end{subfigure}
\begin{subfigure}{0.159\linewidth}
\centering
\includegraphics[scale=0.18]{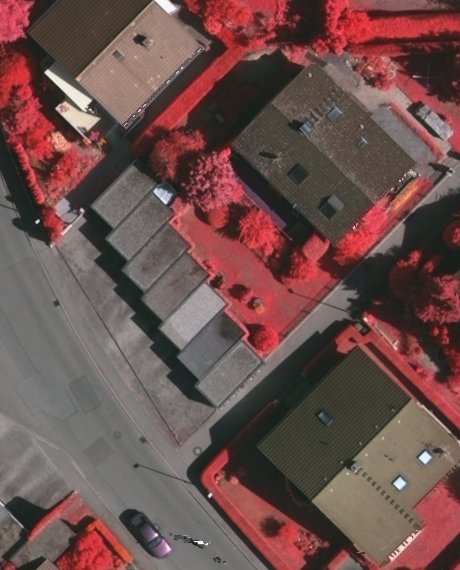}
\end{subfigure}
\begin{subfigure}{0.159\linewidth}
\centering
\includegraphics[scale=0.18]{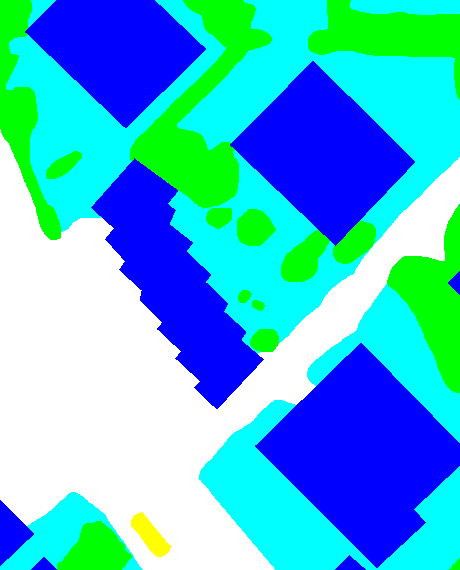}
\end{subfigure}
\begin{subfigure}{0.16\linewidth}
\centering
\includegraphics[scale=0.18]{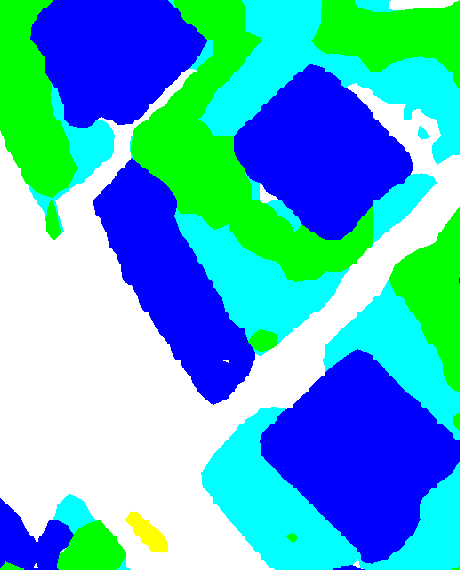}
\end{subfigure}
\begin{subfigure}{0.159\linewidth}
\centering
\includegraphics[scale=0.18]{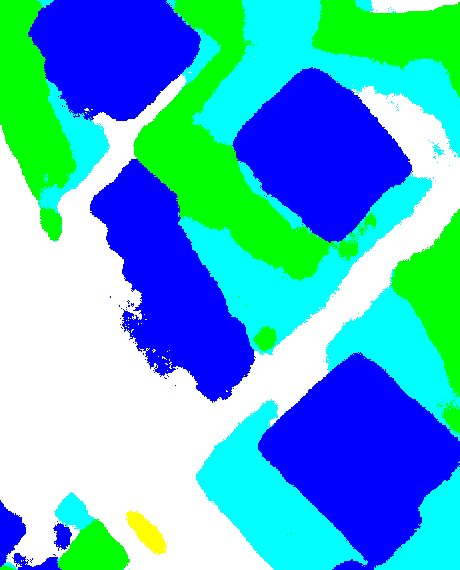}
\end{subfigure}
\begin{subfigure}{0.159\linewidth}
\centering
\includegraphics[scale=0.18]{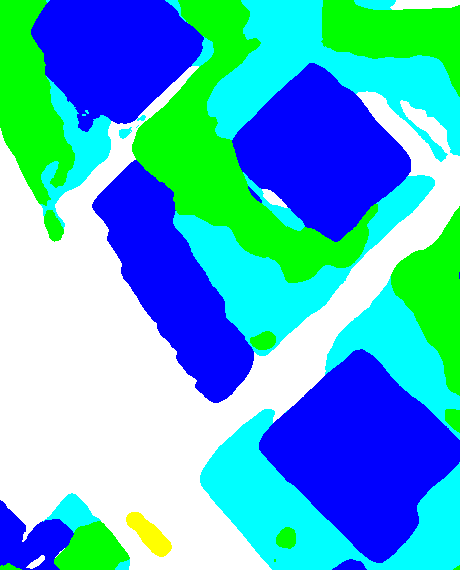}
\end{subfigure}
\begin{subfigure}{0.159\linewidth}
\centering
\includegraphics[scale=0.18]{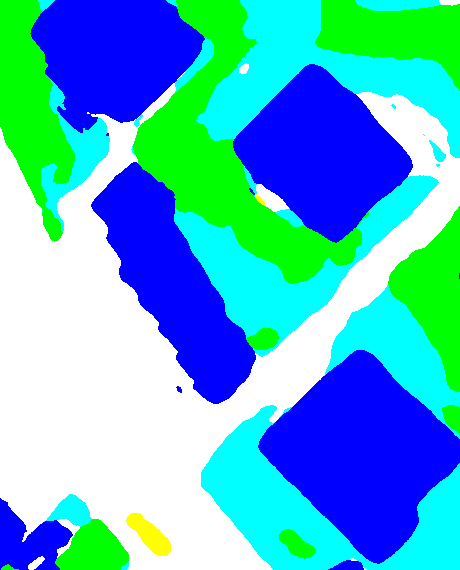}
\end{subfigure}

\begin{subfigure}{0.005\linewidth}
2
\end{subfigure}
\begin{subfigure}{0.159\linewidth}
\centering
\includegraphics[scale=0.18]{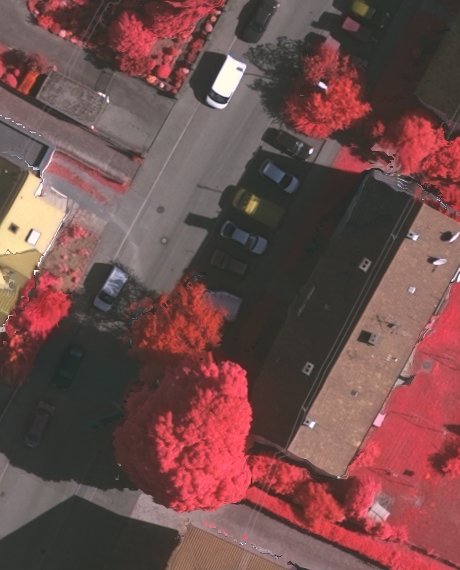}
\end{subfigure}
\begin{subfigure}{0.159\linewidth}
\centering
\includegraphics[scale=0.18]{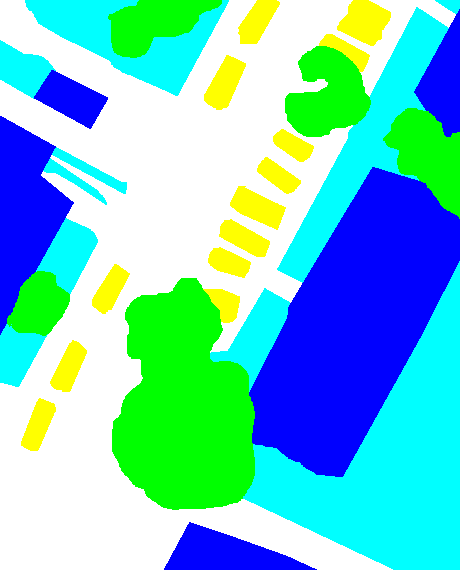}
\end{subfigure}
\begin{subfigure}{0.159\linewidth}
\centering
\includegraphics[scale=0.18]{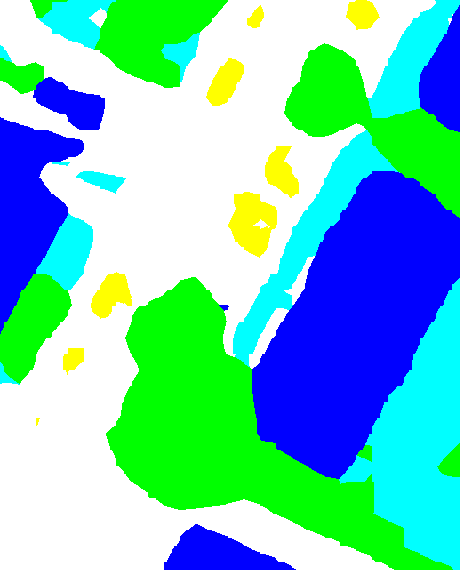}
\end{subfigure}
\begin{subfigure}{0.159\linewidth}
\centering
\includegraphics[scale=0.18]{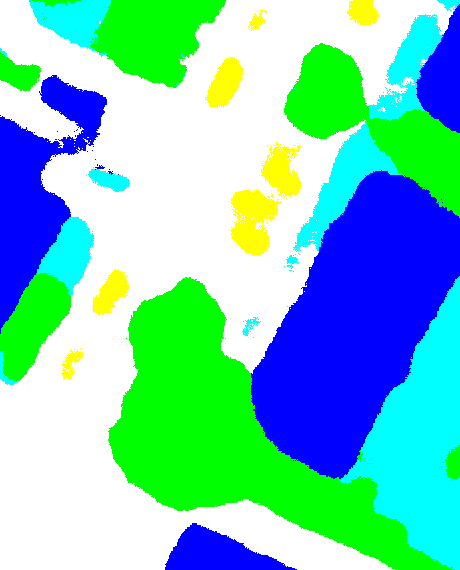}
\end{subfigure}
\begin{subfigure}{0.159\linewidth}
\centering
\includegraphics[scale=0.18]{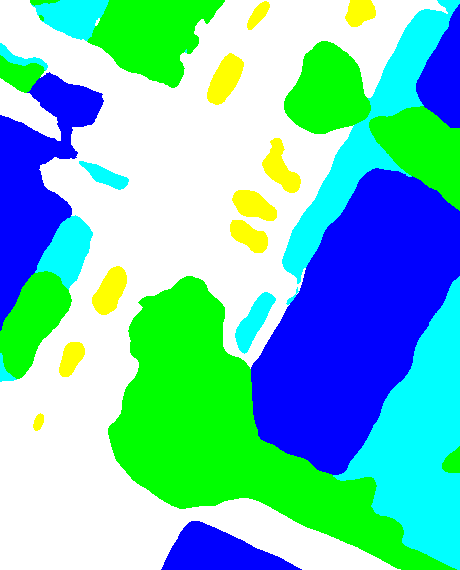}
\end{subfigure}
\begin{subfigure}{0.159\linewidth}
\centering
\includegraphics[scale=0.18]{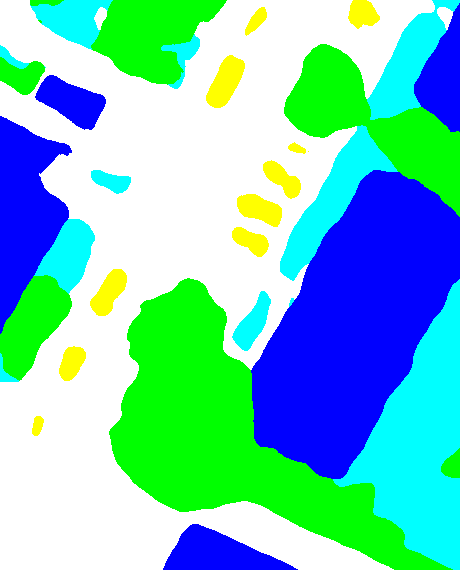}
\end{subfigure}

\begin{subfigure}{0.005\linewidth}
3
\end{subfigure}
\begin{subfigure}{0.159\linewidth}
\centering
\includegraphics[scale=0.72]{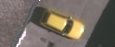}
\end{subfigure}
\begin{subfigure}{0.159\linewidth}
\centering
\includegraphics[scale=0.72]{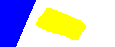}
\end{subfigure}
\begin{subfigure}{0.159\linewidth}
\centering
\includegraphics[scale=0.72]{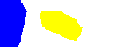}
\end{subfigure}
\begin{subfigure}{0.159\linewidth}
\centering
\includegraphics[scale=0.72]{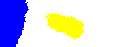}
\end{subfigure}
\begin{subfigure}{0.159\linewidth}
\centering
\includegraphics[scale=0.72]{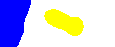}
\end{subfigure}
\begin{subfigure}{0.159\linewidth}
\centering
\includegraphics[scale=0.72]{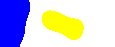}
\end{subfigure}

\begin{subfigure}{0.005\linewidth}
4
\end{subfigure}
\begin{subfigure}{0.159\linewidth}
\centering
\includegraphics[scale=0.12]{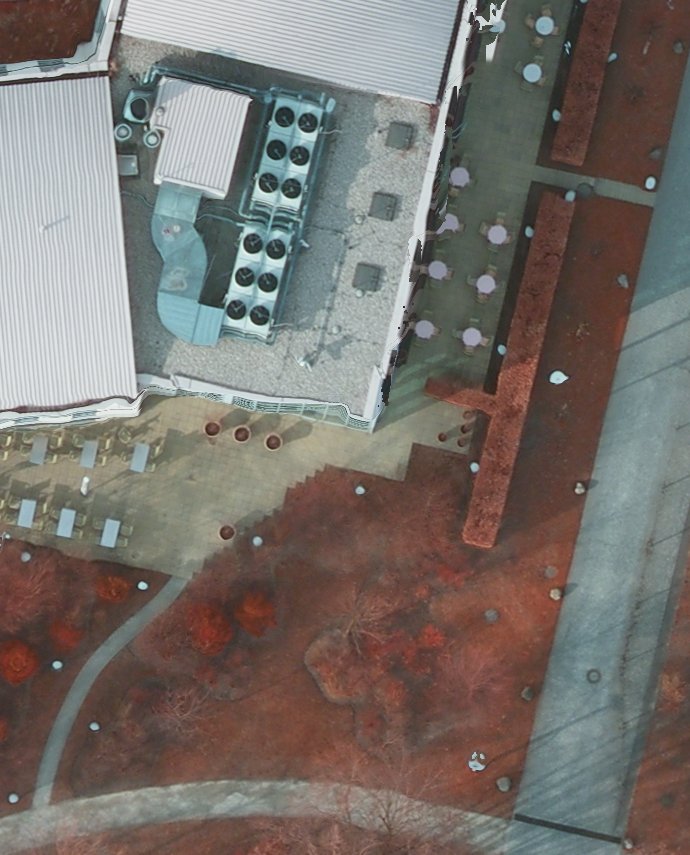}
\end{subfigure}
\begin{subfigure}{0.159\linewidth}
\centering
\includegraphics[scale=0.12]{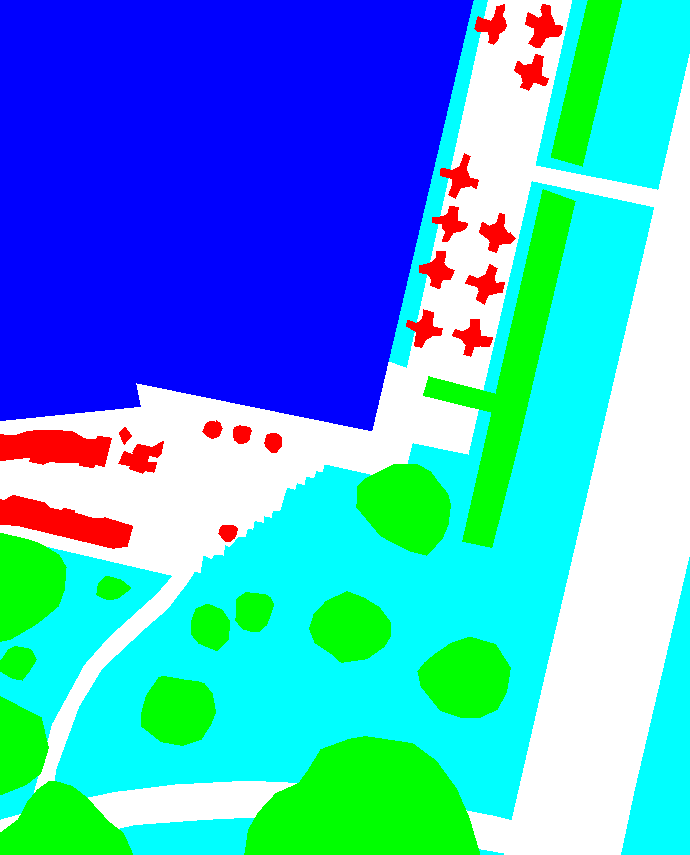}
\end{subfigure}
\begin{subfigure}{0.159\linewidth}
\centering
\includegraphics[scale=0.12]{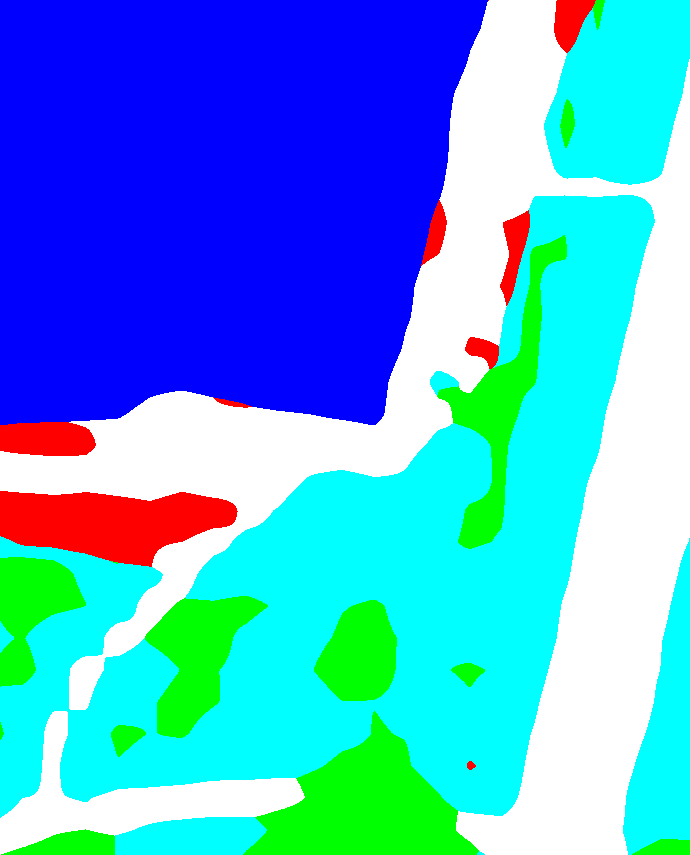}
\end{subfigure}
\begin{subfigure}{0.159\linewidth}
\centering
\includegraphics[scale=0.12]{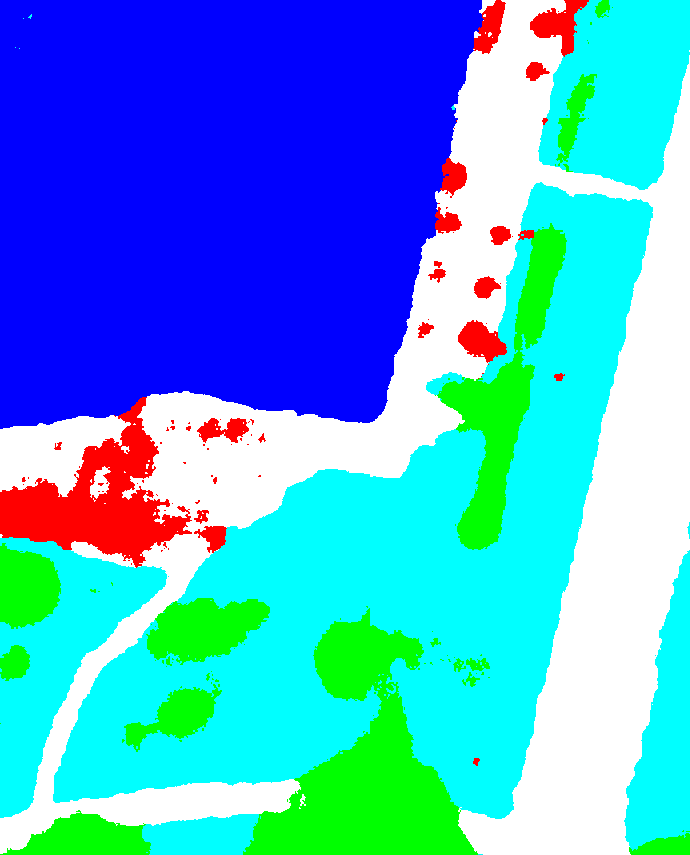}
\end{subfigure}
\begin{subfigure}{0.159\linewidth}
\centering
\includegraphics[scale=0.12]{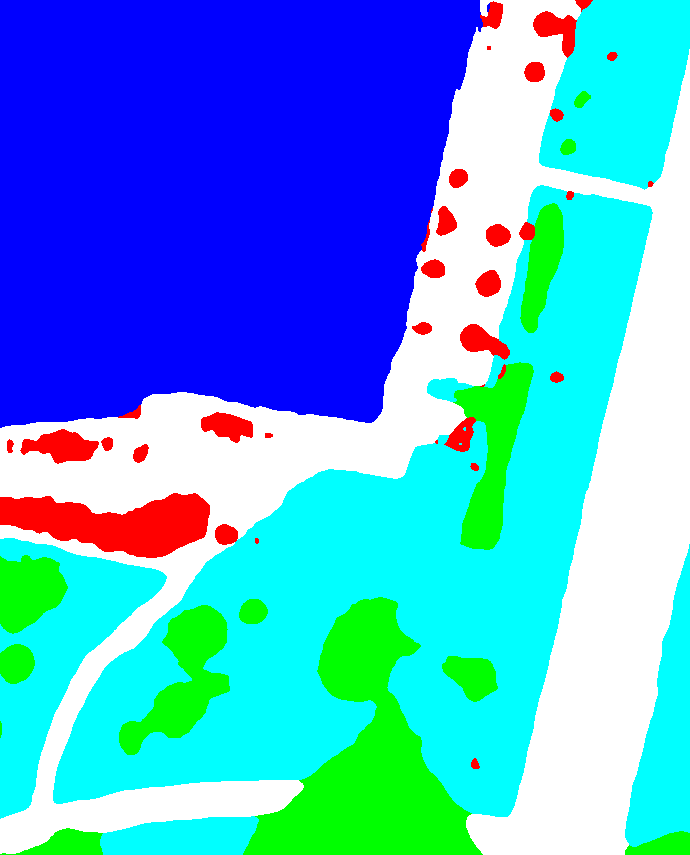}
\end{subfigure}
\begin{subfigure}{0.159\linewidth}
\centering
\includegraphics[scale=0.12]{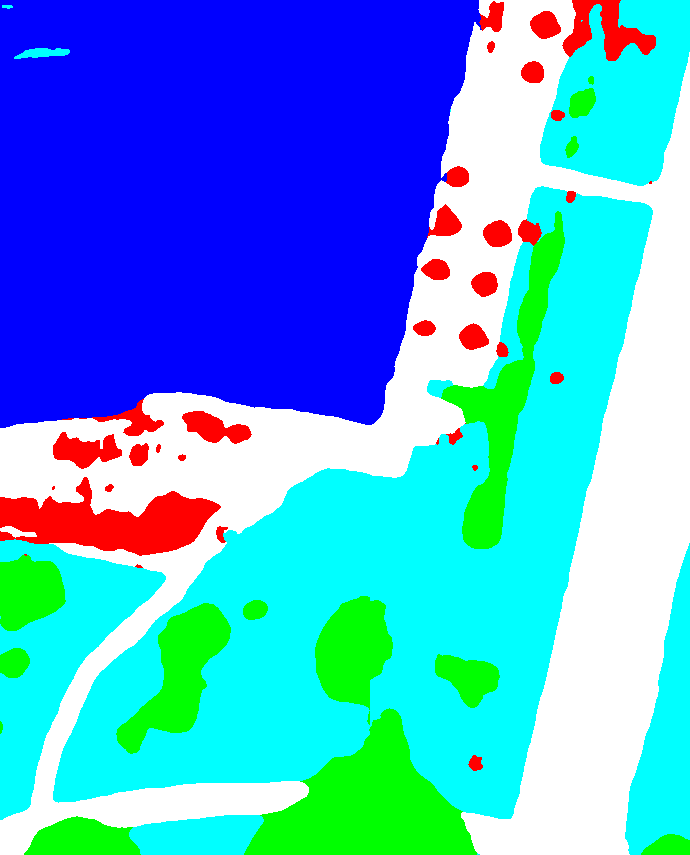}
\end{subfigure}

\begin{subfigure}{0.005\linewidth}
5
\end{subfigure}
\begin{subfigure}{0.159\linewidth}
\centering
\includegraphics[scale=0.09]{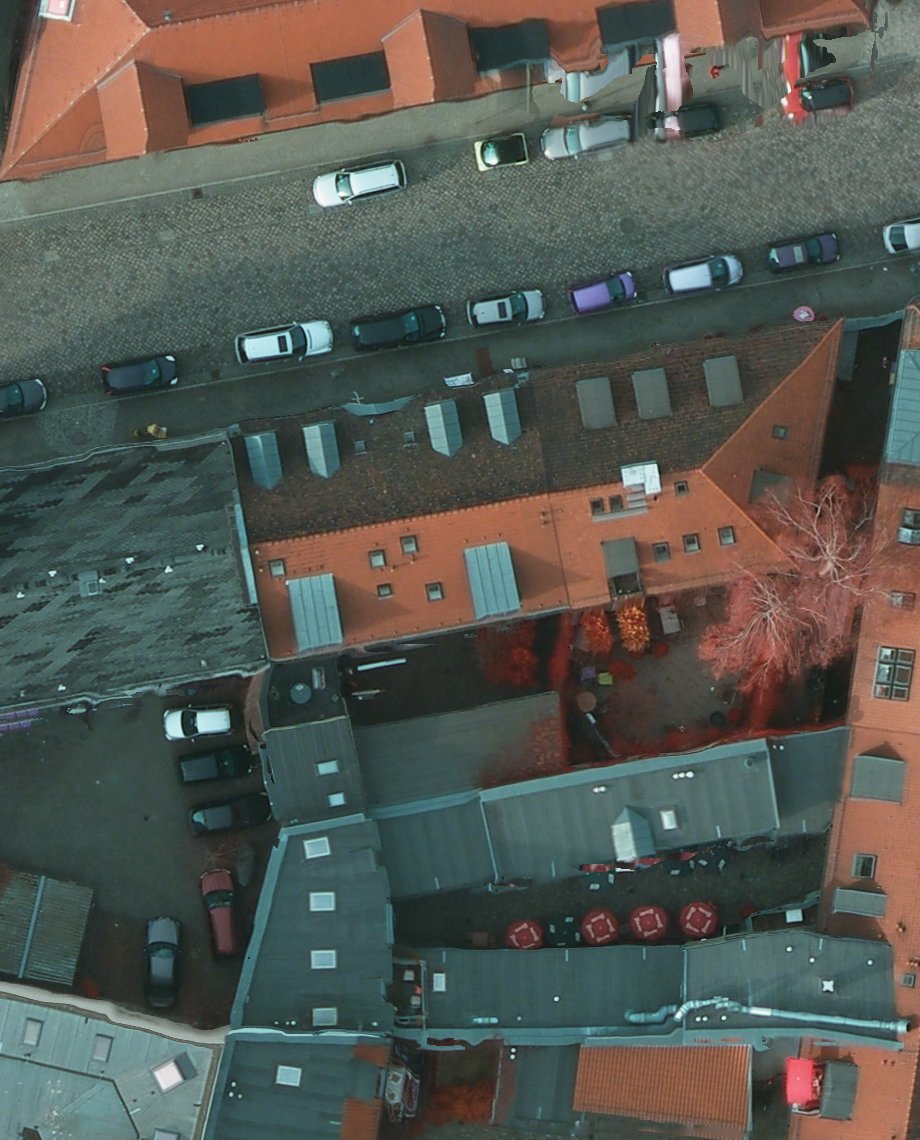}
\end{subfigure}
\begin{subfigure}{0.159\linewidth}
\centering
\includegraphics[scale=0.09]{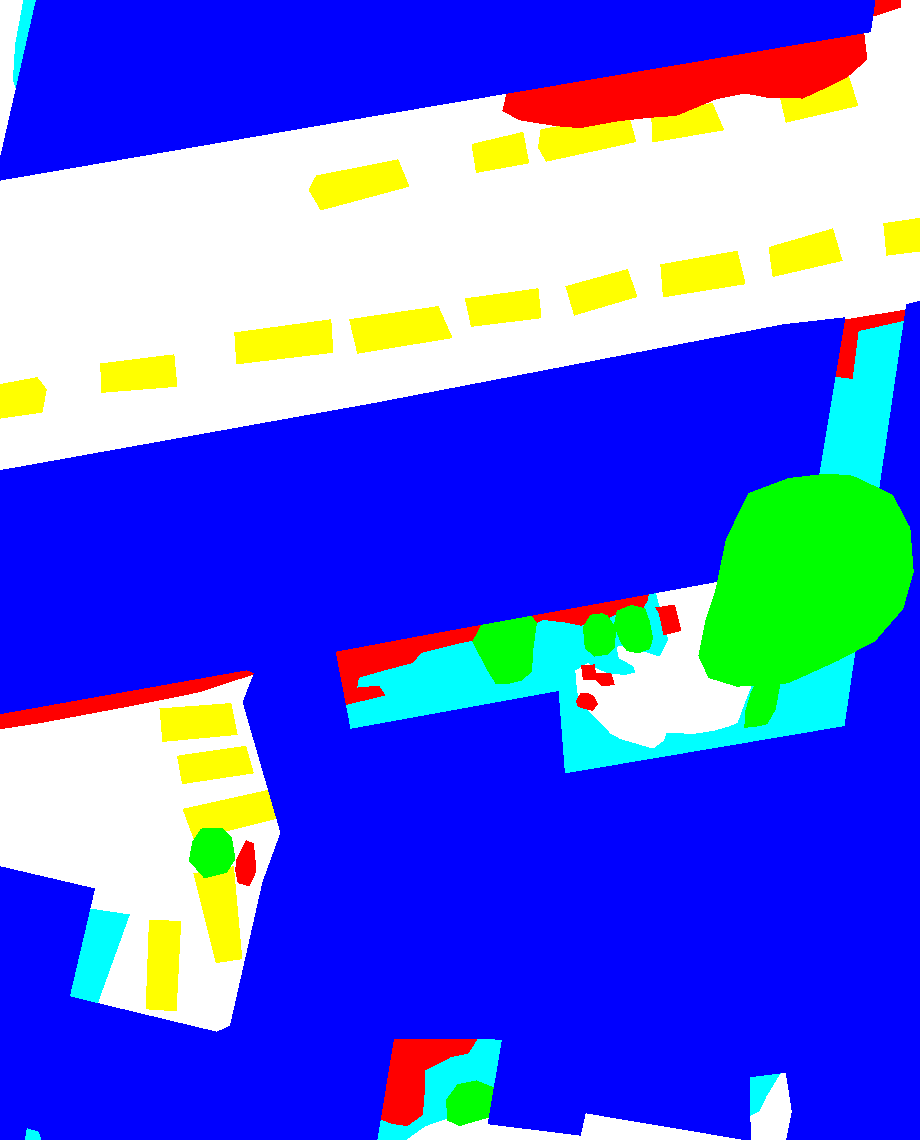}
\end{subfigure}
\begin{subfigure}{0.159\linewidth}
\centering
\includegraphics[scale=0.09]{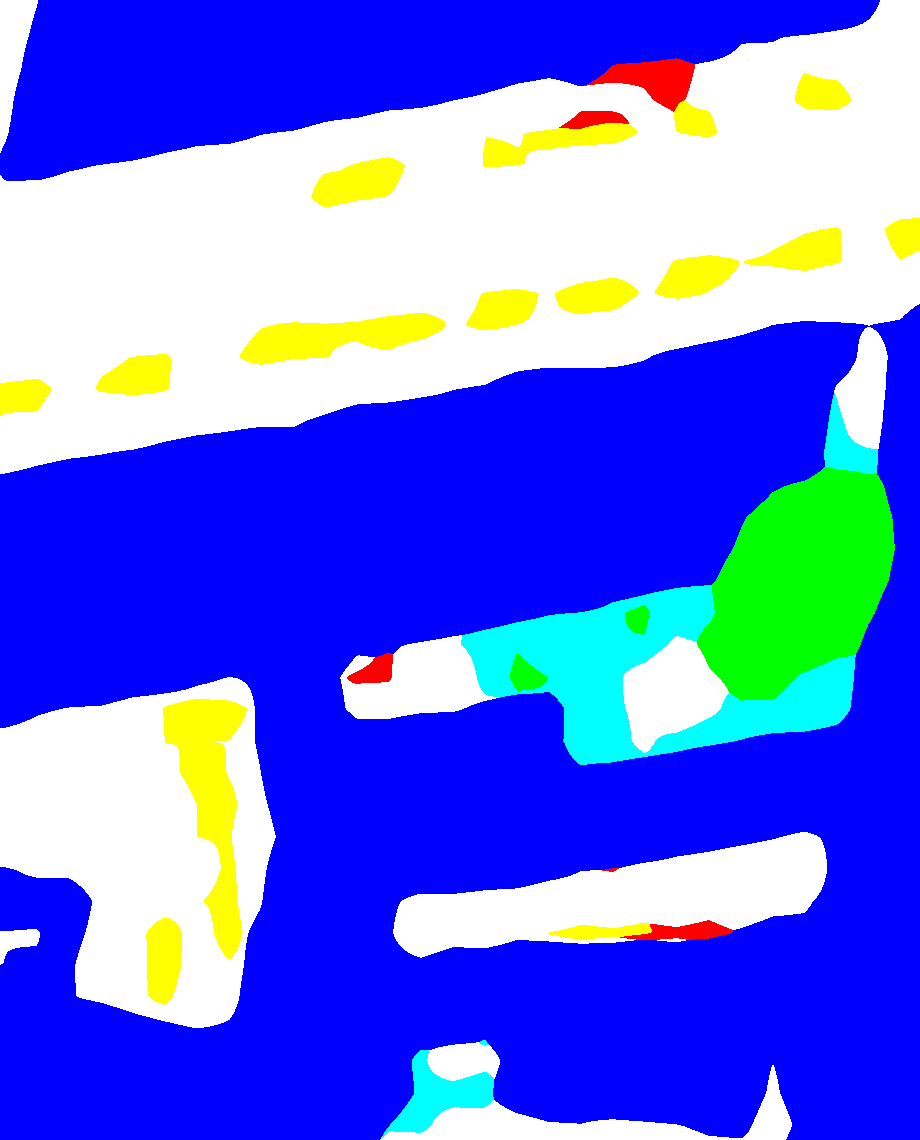}
\end{subfigure}
\begin{subfigure}{0.159\linewidth}
\centering
\includegraphics[scale=0.09]{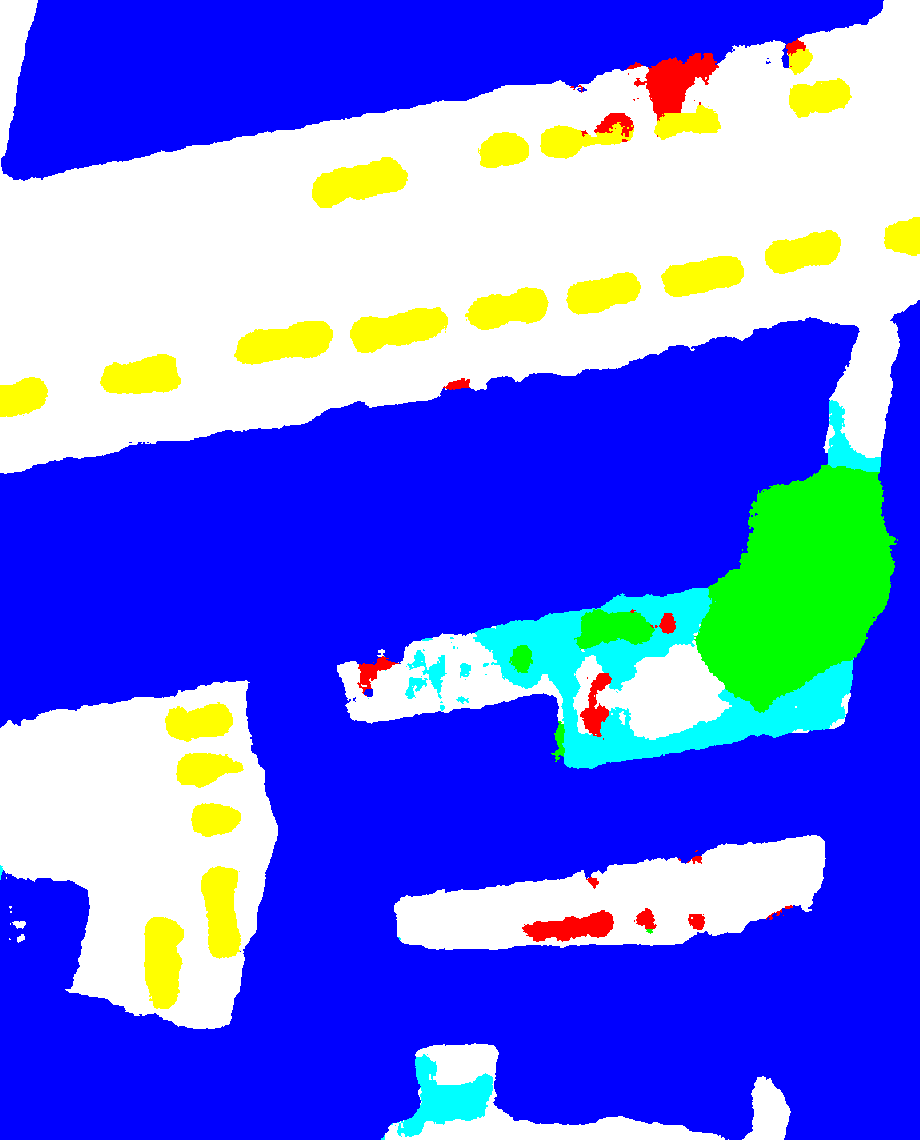}
\end{subfigure}
\begin{subfigure}{0.159\linewidth}
\centering
\includegraphics[scale=0.09]{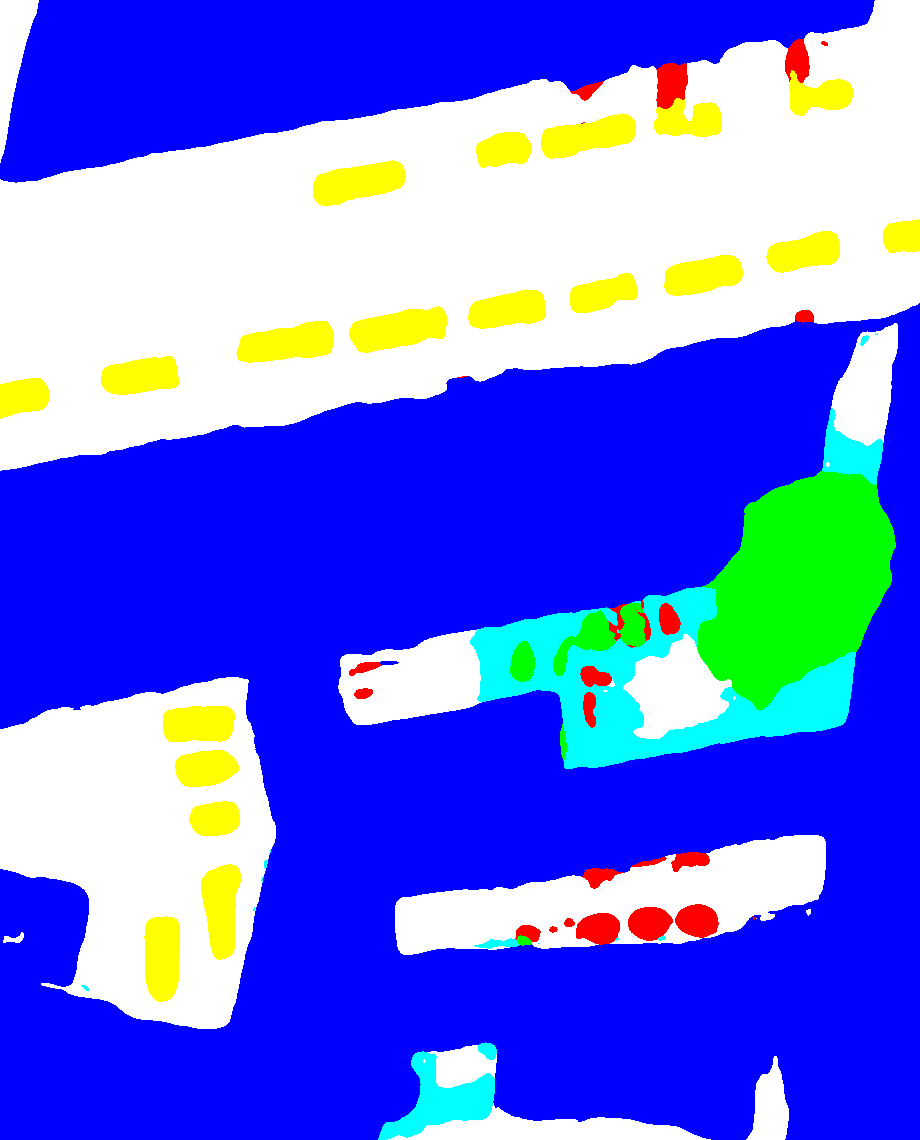}
\end{subfigure}
\begin{subfigure}{0.159\linewidth}
\centering
\includegraphics[scale=0.09]{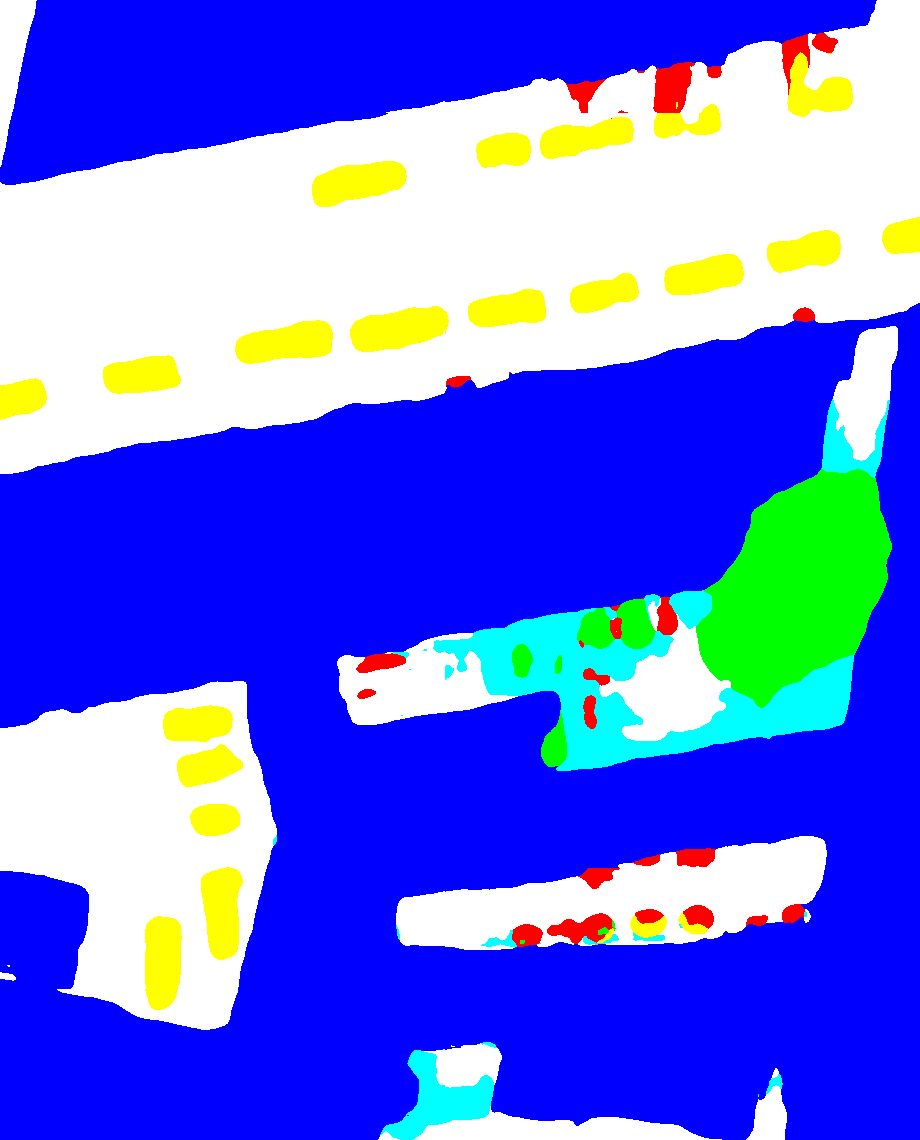}
\end{subfigure}

\begin{subfigure}{0.005\linewidth}
6
\end{subfigure}
\begin{subfigure}{0.159\linewidth}
\centering
\includegraphics[scale=0.2]{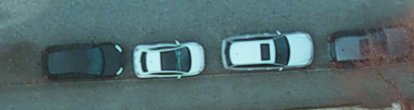}
\end{subfigure}
\begin{subfigure}{0.159\linewidth}
\centering
\includegraphics[scale=0.2]{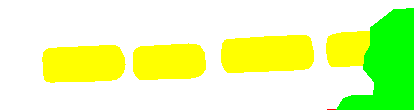}
\end{subfigure}
\begin{subfigure}{0.159\linewidth}
\centering
\includegraphics[scale=0.2]{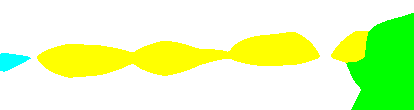}
\end{subfigure}
\begin{subfigure}{0.159\linewidth}
\centering
\includegraphics[scale=0.2]{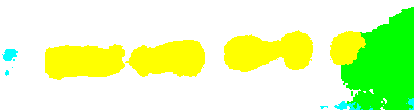}
\end{subfigure}
\begin{subfigure}{0.159\linewidth}
\centering
\includegraphics[scale=0.2]{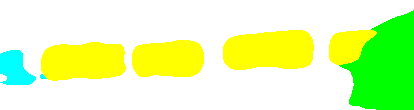}
\end{subfigure}
\begin{subfigure}{0.159\linewidth}
\centering
\includegraphics[scale=0.2]{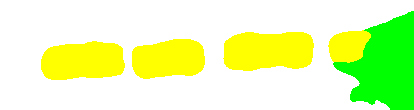}
\end{subfigure}

\caption{Classification of closeups of Vahingen (1--3) and Potsdam (4--6) validation sets. Classes: Impervious surface (white), Building {\color{blue} (blue)}, Low veget. {\color{cyan} (cyan)},  Tree {\color{Green} (green)},  Car {\color{Dandelion} (yellow)}, Clutter {\color{red} (red)}.}
\label{f:visual_frags}
\end{figure*}

\subsection{Visual Results}

\begin{figure*}
\begin{subfigure}{0.24\linewidth}
\centering
Image\end{subfigure}
\begin{subfigure}{0.24\linewidth}
\centering
Deconvolution \cite{volpi}
\end{subfigure}
\begin{subfigure}{0.24\linewidth}
\centering
Dilation \cite{sherrah}
\end{subfigure}
\begin{subfigure}{0.24\linewidth}
\centering
MLP
\end{subfigure}

\begin{subfigure}{0.24\linewidth}
\centering
\includegraphics[scale=0.16,angle=90]{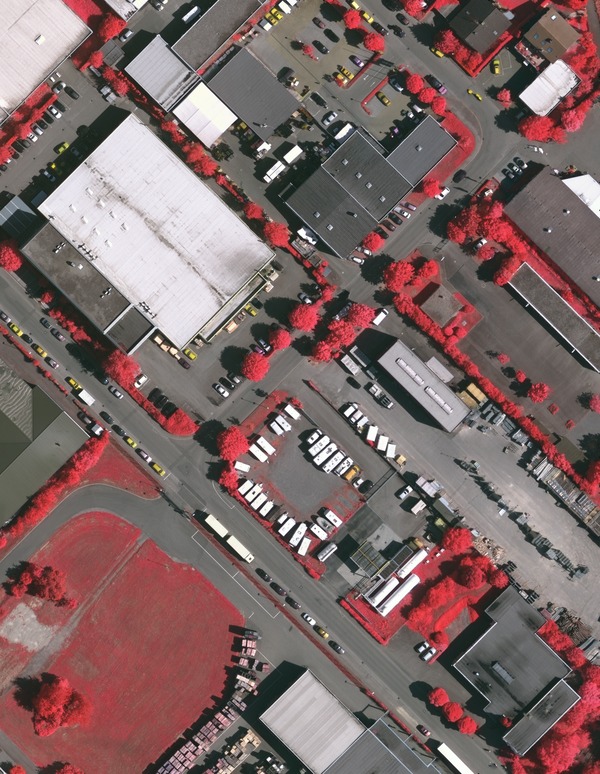}
\end{subfigure}
\begin{subfigure}{0.24\linewidth}
\centering
\includegraphics[scale=0.16,angle=90]{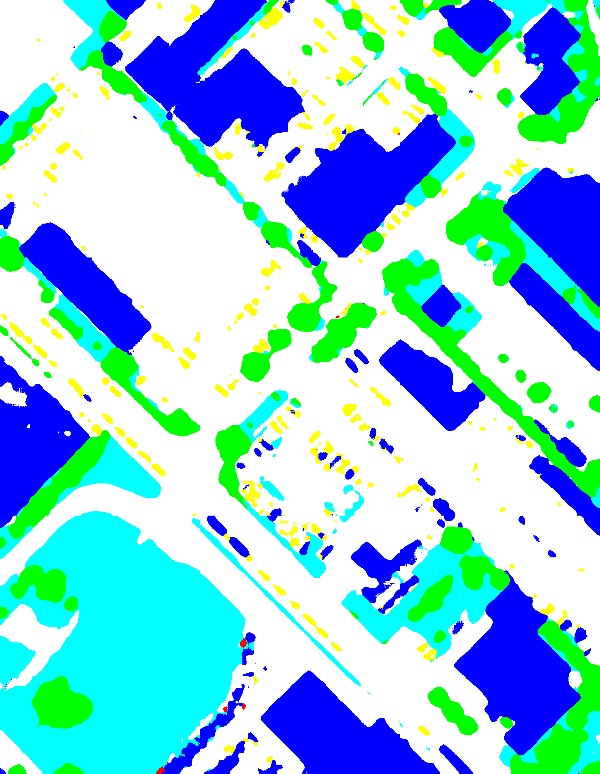}
\end{subfigure}
\begin{subfigure}{0.24\linewidth}
\centering
\includegraphics[scale=0.16,angle=90]{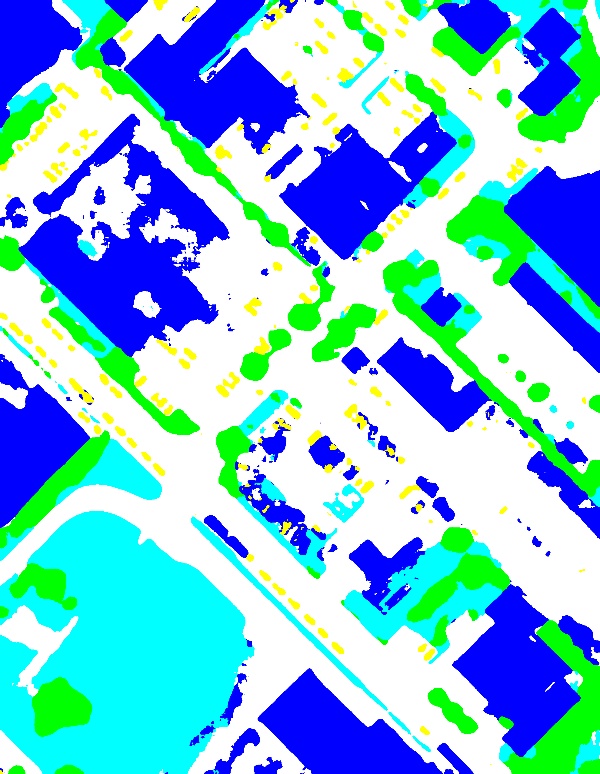}
\end{subfigure}
\begin{subfigure}{0.24\linewidth}
\centering
\includegraphics[scale=0.16,angle=90]{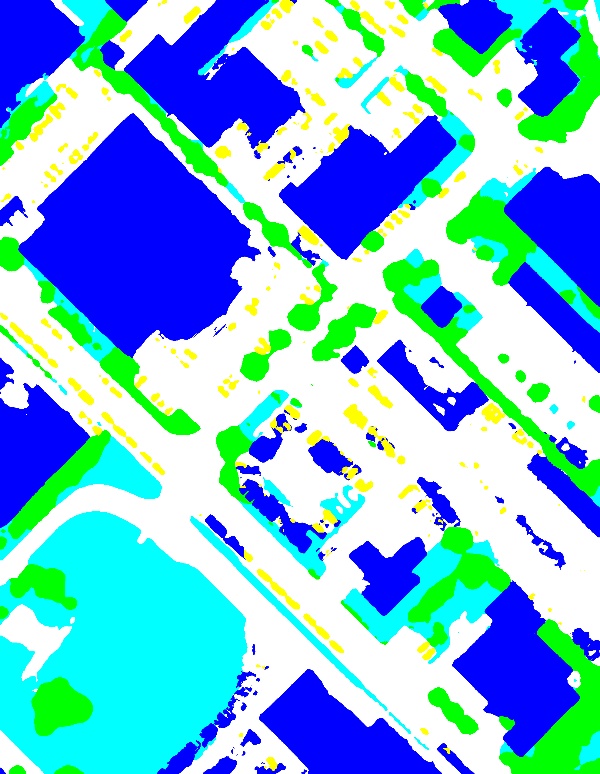}
\end{subfigure}

\begin{subfigure}{0.24\linewidth}
\centering
\includegraphics[scale=0.207]{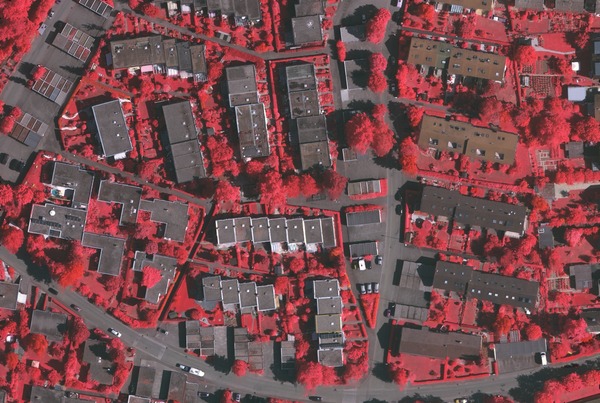}
\end{subfigure}
\begin{subfigure}{0.24\linewidth}
\centering
\includegraphics[scale=0.207]{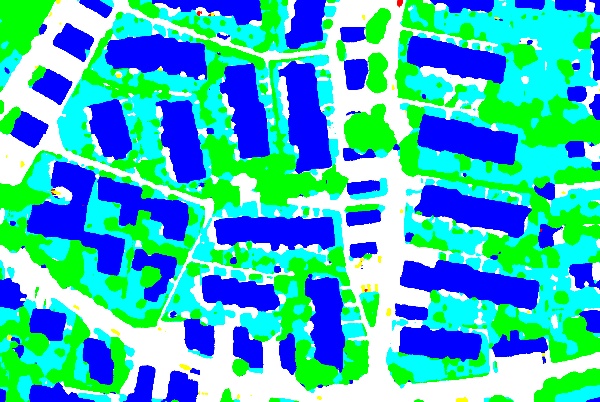}
\end{subfigure}
\begin{subfigure}{0.24\linewidth}
\centering
\includegraphics[scale=0.207]{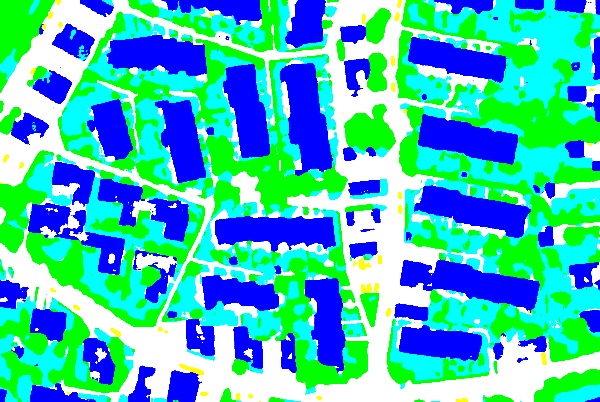}
\end{subfigure}
\begin{subfigure}{0.24\linewidth}
\centering
\includegraphics[scale=0.207]{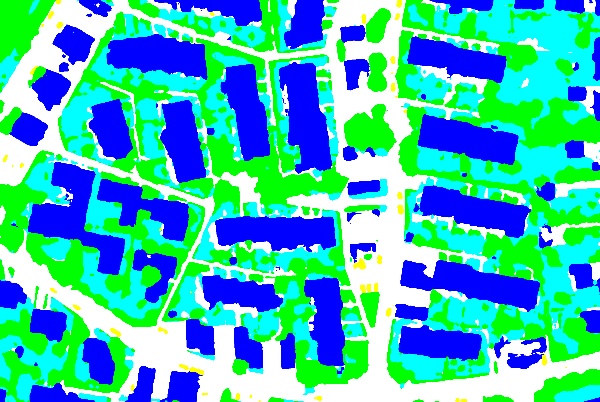}
\end{subfigure}
\caption{Classification of entire tiles of the Vaihingen test set.}
\label{f:entire_tile}
\end{figure*}

We include visual comparisons on closeups of classified images of both datasets in Fig.~\ref{f:visual_frags}. As expected, the base FCN tends to output ``blobby'' objects, while the other methods provide sharper results. This is particularly noticeable for the cars of Rows 2, 5 and 6, and for the thin road at the lower left corner of Row~4. We also observe that 
the incorporation of 
reasoning at lower resolutions allows the derived networks to discover small objects that are otherwise lost. This is particularly noticeable in the 4th row, where there 
is a set 
of small round/cross-shaped objects of the \emph{clutter} class (in red) that are omitted or grouped together by the base FCN. 

The unpooling technique seems to be prone to outputting artifacts. These are often very small in size, even isolated pixels. This is well observed for example for the car of Row~3. This effect  could be a natural consequence of the max unpooling mechanism, as depicted in Fig.~\ref{f:unpooling}, which upsamples into sparse matrices and delegates the task of reconstructing a smoother output to the deconvolutional layers.

At first sight it is more challenging to visually assess why MLP beats the skip network in almost every case in the numerical evaluation. Taking a closer look we can however observe that boundaries tend to be more accurate at a fine level in the case of MLP. For example, 
the ``staircase'' shape of one of the buildings in Row 1 is 
noticeably
better outlined by the MLP network. 

We can also observe that the ground truth itself is often not very precise. For example, the car in Row 3 does not seem to be labeled accurately, hence it is  difficult to imagine that a network would learn to finely label that class. 
In Row 5, an entire lightwell between buildings has apparently  been omitted in the ground truth (labeled as part of the building), yet recognized as an impervious surface by the CNNs.

The general recognition capabilities of CNNs can also be well appreciated in these fragments. For example, in Row 4, while there are tiny round objects both on the roof of the building and outside the building, CNNs correctly label as \emph{building} the ones on the roof and as \emph{clutter} the other ones. 

In Fig.~\ref{f:entire_tile} we show the classification of entire tiles of the Vaihingen set, obtained from the test set submissions. We include the \emph{deconvolution}~\cite{volpi} and \emph{dilation}~\cite{sherrah} network results, together with our MLP. We can see, for instance, that a large white building in the first image is recognized by MLP but misclassified  or only partially recovered by the other methods. In the second tile, the Dilation method outputs some holes in the buildings which are not present in the MLP results. A better combination of the information coming from different resolutions might explain why MLP  successfully recognizes that these entire surfaces do belong to the same object.

\subsection{Running Times}
\label{s:runningtimes}

Table~\ref{t:exectime} reports the running times for training and testing on both datasets. The training time of the architectures derived from the base FCN comprises the time to pretrain the base FCN first and the time to then train the whole system altogether (see details in Sec.~\ref{s:exp:training}). The architectures were implemented using Caffe~\cite{caffe} and run on an Intel I7 CPU @ 2.7Ghz
with a Quadro K3100M GPU (4 GB RAM). We also add for comparison the results reported by the author of the Dilation network~\cite{sherrah}, run on a larger 12 GB RAM GPU. To classify large images we crop them into tiles with as much overlap as the amount of padding in the network, to avoid tile border effects. 

As reported in the table, the unpooling, skip and MLP networks introduce an overhead to the base FCN. MLP is the slowest of the derived networks, followed by the unpooling and skip networks. 
MLP, which provides the highest accuracy, classifies the entire Vaihingen validation set in about 30 seconds and the Postdam validation set in 2 minutes. This is much faster than the dilation network. Incorporating the principle of Fig.~\ref{f:fullnotfullres} allows us to better allocate computational resources, not spending too much time and space in conducting a high-resolution analysis where it is not needed, boosting accuracy and performance.

\section{Concluding Remarks}
\label{s:concl}

Convolutional neural networks (CNNs) are becoming the leading choice for high-resolution semantic labeling. The biggest concern with this technique is the spatial coarseness of the outputs. Most of the work has moderately modified or post-processed well-known CNN architectures in order to counteract this issue. 
 We decided, however, to rethink CNNs from a semantic labeling perspective.

For this purpose, we first analyzed different families of semantic labeling CNN prototypes. This analysis bears some similarity with the reasoning that gave birth to CNNs themselves: we study which relevant constraints can be imposed in the architecture by construction, reducing the number of parameters and improving the optimization. We observed that existing networks often spend efforts in learning invariances that could be otherwise guaranteed, and reason at a high resolution even when it is not needed. While previous methods are 
already 
competitive, we can devise more optimal approaches. 

We derived a model in which spatial features are learned at multiple resolutions (and thus different levels of detail) and a specific CNN module learns how to combine them. In our experiments on aerial imagery, such a model proved to be more effective than the other approaches to conduct high-resolution labeling. It provides a better accuracy with low computational requirements, leading to a win-win situation. Some of the outperformed methods are in fact significantly more complex than our approach, proving once again that striving for simplicity is often the way to go when using CNN architectures. 

We hope that our architectural prototype will be used as a basis for semantic labeling networks.
Our future plan is to create a large-scale aerial image dataset, covering dissimilar areas of the earth, on which to conduct semantic labeling with convolutional neural networks.

\section*{Acknowledgment}

The authors would like to thank CNES for initializing and funding this study.

\ifCLASSOPTIONcaptionsoff
  \newpage
\fi

\bibliographystyle{IEEEbib}
\bibliography{biblio}

\vspace{15pt}
\footnotesize \noindent \textbf{Emmanuel Maggiori} (S'15) received the Engineering degree in computer science from Central Buenos Aires Province National University (UNCPBA), Tandil, Argentina, in 2014. The same year he joined AYIN and STARS teams at Inria Sophia Antipolis-M\'editerran\'ee as a research intern in the field of remote sensing image processing. Since 2015, he has been working on his Ph.D.~within TITANE team, studying machine learning techniques for large-scale processing of satellite imagery. 

\vspace{15pt}
\footnotesize \noindent \textbf{Yuliya Tarabalka} (S'08--M'10) received the B.S. degree in computer science from Ternopil Ivan Pul'uj State Technical University, Ukraine, in 2005 and the M.Sc. degree in signal and image processing from the Grenoble Institute of Technology (INPG), France, in 2007. She received a joint Ph.D. degree in signal and image processing from INPG and in electrical engineering from the University of Iceland, in 2010.

From July 2007 to January 2008, she was a researcher with the Norwegian Defence Research Establishment, Norway. From September 2010 to December 2011, she was a postdoctoral research fellow with the Computational and Information Sciences and Technology Office, NASA Goddard Space Flight Center, Greenbelt, MD. From January to August 2012 she was a postdoctoral research fellow with the French Space Agency (CNES) and Inria Sophia Antipolis-M\'editerran\'ee, France. She is currently a researcher with the TITANE team of Inria Sophia Antipolis-M\'editerran\'ee. Her research interests are in the areas of image processing, pattern recognition and development of efficient algorithms. She is Member of the IEEE Society.

\vspace{15pt}
\footnotesize \noindent \textbf{Guillaume Charpiat}
received the B.S. degree in mathematics and physics from the \'Ecole Normale
Sup\'erieure (ENS) at Paris, France, the M.Sc. degree
in computer vision and machine learning, and theoretical physics from ENS at Cachan, France, and the Ph.D. degree in computer science at ENS in 2006.
His Ph.D. thesis was on the distance-based shape
statistics for image segmentation with priors.

He was with the Max-Planck Institute for Biological Cybernetics (T\"ubingen, Germany), where he was involved in medical imaging (MR-based PET prediction) and automatic image colorization. He was a researcher with Inria Sophia Antipolis-M\'editerran\'ee, Valbonne, France, where he was involved in image segmentation and optimization techniques. He is currently a Researcher with the TAO team, Inria Saclay, Palaiseau, France, where he is involved in machine learning, in particular on building a theoretical background for neural networks.

\vspace{15pt}
\footnotesize \noindent \textbf{Pierre Alliez} is Senior Researcher and team leader at Inria Sophia-Antipolis - M\'editerran\'ee, Valbonne, France. He has authored scientific publications and several book chapters on mesh compression, surface reconstruction, mesh generation, surface remeshing and mesh parameterization.

 Dr. Alliez was a recipient of the EUROGRAPHICS Young Researcher Award in 2005 for his contributions to computer graphics and geometry processing and a Starting Grant from the European Research Council on Robust Geometry Processing in 2011. He was the co-chair of the Symposium on Geometry Processing in 2008, Pacific Graphics in 2010 and Geometric Modeling and Processing 2014. He is currently an Associate Editor of the Computational Geometry Algorithms Library and an Associate Editor of the ACM Transactions on Graphics.

\end{document}